\documentclass{article}

\usepackage[preprint]{neurips_2018}

\usepackage{color}
\usepackage[utf8]{inputenc} % allow utf-8 input
\usepackage[T1]{fontenc}    % use 8-bit T1 fonts
\usepackage[draft]{hyperref}     % hyperlinks
\usepackage{url}            % simple URL typesetting
\usepackage{booktabs}       % professional-quality tables
\usepackage{amsfonts}       % blackboard math symbols
\usepackage{nicefrac}       % compact symbols for 1/2, etc.
\usepackage{microtype}      % microtypography
\usepackage{bm}
\usepackage{amsmath}
\usepackage{graphicx}
\usepackage{theorem}

\title{Wavelet regression and additive models for irregularly spaced data}

\newcommand{\bs}[1]{\boldsymbol{#1}}

\newcommand{\wt}[1]{\widetilde{#1}}
\newcommand{\wh}[1]{\widehat{#1}}

\newcommand{\e}{\varepsilon}
\newcommand{\name}{\texttt{waveMesh}}

\newtheorem{theorem}{Theorem}
\newtheorem{definition}{Definition}

\author{
  Asad Haris\thanks{Mailing address: Box 357232, University of Washington, Seattle, WA 98195-7232} \\
  Department of Biostatistics\\
  University of Washington\\
  Seattle, WA 98195 \\
  \texttt{aharis@uw.edu} \\
   \And
   Noah Simon \\
Department of Biostatistics\\
University of Washington\\
Seattle, WA 98195 \\
\texttt{nrsimon@uw.edu} \\
  \AND
  Ali Shojaie \\
  Department of Biostatistics\\
  University of Washington\\
  Seattle, WA 98195 \\
  \texttt{ashojaie@uw.edu} \\
}

\begin{document}

\maketitle

% !TeX root = ../Manuscript.tex

\begin{abstract}
	We present a novel approach for nonparametric regression using wavelet basis functions. Our proposal, \name, can be applied to non-equispaced data with sample size not necessarily a power of 2. We develop an efficient proximal gradient descent algorithm for computing the estimator and establish adaptive minimax convergence rates. The main appeal of our approach is that it naturally extends to additive and sparse additive models for a potentially large number of covariates. We prove minimax optimal convergence rates under a weak compatibility condition for sparse additive models. The compatibility condition holds when we have a small number of covariates. Additionally, we establish convergence rates for when the condition is not met. We complement our theoretical results with empirical studies comparing \name\ to existing methods.
\end{abstract}

% !TeX root = ../Manuscript.tex

\section{Introduction}
\label{sec:introduction}

We consider the canonical task of estimating a regression function, $f$, from observations $\{(\bm{x}_i, y_i): i = 1,\ldots, n\}$, with $\bm{x}_i\in[0,1]^p$, $y_i\in\mathbb{R}$ and $y_i = f(\bm{x}_i) + \e_i$ ($i=1,\ldots, n$), where $\e_i$ are independent, mean 0, sub-Gaussian random variables. 
A popular approach for estimating $f$ is to use linear combinations of a pre-specified set of \emph{basis functions}, e.g., polynomials, splines~\citep{wahba1990spline}, wavelets~\citep{daubechies1992ten}, or other systems \citep{chentsov1962evaluation}. The weights, or coefficients, in such a linear combination are often determined using some form of penalized regression. In this paper, we focus on estimators that use \emph{wavelets}.
Wavelet-based estimators have compelling theoretical properties. However, a number of issues have limited their  adaptation in many non-parametric applications. %We discuss these issues in the reminder of this section. 
The approach proposed in this paper overcomes these issues. 
Throughout the paper, we assume basic knowledge of wavelet methods though some key points will be reviewed. For a detailed introduction to wavelets, see books by \cite{daubechies1992ten, percival2006wavelet, vidakovic2009statistical, nason2010wavelet, ogden2012essential}. 

Wavelets are a system of orthonormal basis functions for $L^2([0,1])$.
Wavelets are popular for representing functions because they allow \emph{time and frequency localization}~\citep{daubechies1990wavelet} as opposed to, say, Fourier bases, which allow only frequency localization. Additionally, wavelet-based methods are computationally efficient. The main ingredient of wavelet regression is the discrete wavelet transform (DWT) and its inverse (IDWT) which can be computed in $O(n)$ operations~\citep{mallat1989theory}. Unfortunately, traditional wavelet methods require stringent conditions on the data, specifically that  $x_i = i/n$ with $n = 2^J$ for some integer $J$. This is not a problem in many signal processing applications with regularly sampled signals; however, in general non-parametric regression, this condition will rarely be satisfied. A simple solution for general data types is to ignore irregular spacing of data~\citep{cai1999wavelet, sardy1999wavelet} and/or artificially extend the signal such that $n = 2^J$~\citep[Ch. 8]{strang1996wavelets}. Other solutions include transformations~\citep{cai1998wavelet,pensky2001non} or interpolation~\citep{hall1997interpolation, kovac2000extending, antoniadis2001regularization} of the data to a regular grid of size $2^J$. The literature on univariate wavelet methods is quite extensive and cannot be adequately discussed within this manuscript. In contrast, the literature on wavelet methods for multiple covariates is rather limited, particularly when the number of covariates is large. 
%In this paper, we give a simple proposal that effectively extends wavelet-based methods to non-parametric modeling with a potentially large number of covariates.
%More recent techniques include block thresholding methods~\citep{chicken2007nonparametric,chesneau2007wavelet}, wavelet reproducing kernel Hilbert spaces~\citep{amato2006wavelet}, and adaptive lifting~\citep{nunes2006adaptive}. 

For the multivariate settings with $\bs{x}_i\in [0,1]^p$ for $p\ge 2$, we consider estimating an additive model, i.e., $\wh{f}\left(\bs{x}_i\right) = \sum_j \wh{f}_j\left(x_{ij}\right)$. Additive models naturally extend linear models to capture non-linear conditional relationships, while retaining some interpretability; they also do not suffer from the \emph{curse of dimensionality}. Despite these benefits, wavelet-based additive models have received limited attention. This is most likely because data with multiple covariates are rarely available on a regular grid of size $n = 2^J$. \cite{sardy2004amlet} fit additive wavelet models by treating the data as if regularly spaced; however, they do not discuss the case when $n$ is not a power of $2$. A number of proposals transform the data to a regular grid~\citep{amato2001adaptive, zhang2003wavelet, grez2018empirical}. However, to do this, the density of the covariates must be estimated, which unnecessarily invokes the curse of dimensionality. In addition, to the best of our knowledge, there are no wavelet-based methods for fitting additive models in high dimensions (when $p>n$) that induce sparsity, i.e., for many $j$, give a solution with $\wh{f}_j \equiv 0$. 

%In the high-dimensional setting, particularly when $p>n$, a common is assumption is \emph{sparsity}, that is, we assume $f_j\equiv 0$ for many values of $j$. There have been numerous proposals for fitting sparse additive models in the literature (see e.g., \cite{meier2009high,ravikumar2009sparse,lou2016sparse}) however to the best of our knowledge this has not been done for wavelet based methods. 

In this paper, we give a simple proposal that effectively extends wavelet-based methods to non-parametric modeling with a potentially large number of covariates.
We present an interpolation-based approach for dealing with irregularly spaced data when $n$ is not necessarily a power of $2$. However, unlike existing interpolation methods, we do not transform the raw data $(\bm{x}_i, y_i)$. As a result, our method naturally extends to additive and sparse additive models. We also propose a penalized estimation framework to induce sparsity in high dimensions. We develop a proximal gradient descent method for computation of our estimator, which leverages fast algorithms for DWT and sparse matrix multiplication. Furthermore, we establish adaptive minimax convergence rates (up to a $\log n$ factor) similar to that of existing wavelet methods for regularly spaced data. We also establish convergence rates for our (sparse) additive proposal for a potentially large number of covariates. We discuss an extension of our proposal to general convex loss functions, and a weighted variation of our penalty which exhibits improved performance.  

%The rest of this paper is organized as follows. 
In Section~\ref{sec:methods} we present our univariate, additive and sparse additive proposals. The univariate case ($p=1$) is mainly presented to motivate our proposal. We also present our main algorithm for computing the estimator. We establish convergence rates of our estimators in Section~\ref{sec:theory}, and present empirical studies in Section~\ref{sec:simulations}. Concluding remarks are given in Section~\ref{sec:conclusion}.

% !TeX root = ../Manuscript.tex

\section{Methodology}
\label{sec:methods}
\subsection{Short background on wavelets}
\label{sec:background}
We begin with a quick review of wavelet methods for nonparametric regression covering 3 main ingredients: (1) wavelet basis functions, (2) the discrete wavelet transform (DWT) and, (3) shrinkage.

First, \emph{wavelets} are a system of orthonormal basis functions for $L^2([0,1])$ or $L^2(\mathbb{R})$. The bases are generated by translations and dilations of special functions $\phi(\cdot)$ and $\psi(\cdot)$ called the \emph{father} and \emph{mother} wavelet, respectively. In greater detail, for any $j_0\ge 0$, a function $f\in L^2([0,1])$ can be written as
\begin{equation}
\label{eqn:waveletExpansion}
f(x) = \sum_{k=0}^{2^{j_0}-1}\alpha_{j_0k}\phi_{j_0k}(x) + \sum_{j=j_0}^{\infty}\sum_{k=0}^{2^j-1}\beta_{jk}\psi_{jk}(x),
\end{equation}
where
\[
\phi_{jk}(x) = 2^{j/2}\phi(2^jx-k), \quad \psi_{jk}(x) = 2^{j/2}\psi(2^jx-k).
\]
The coefficients $\alpha_{j_0k}$ and $\beta_{jk}$ are called the father and mother wavelet coefficients, respectively. The index $j$ is called the \emph{resolution level} and $j_0$ is the \emph{minimum resolution level}. Different choices of $\phi$ and $\psi$ generate various wavelet families. Popular choices are Daubechies~\citep{daubechies1988orthonormal}, Coiflets~\citep{daubechies1993orthonormal}, Meyer wavelets~\citep{meyer1985principe}, and Spline wavelets~\citep{chui1992introduction}; for an overview of wavelet families, see \cite{ogden2012essential}. 
Often functions with a truncated basis expansion are considered, i.e., functions of the form $f(x) = \sum_{k=0}^{2^{j_0}-1}\alpha_{j_0k}\phi_{j_0k}(x) + \sum_{j=j_0}^{J}\sum_{k=0}^{2^j-1}\beta_{jk}\psi_{jk}(x)$, for some $J$. For regular data with $x_i = i/n\ (i=1,\ldots,n)$ and $n = 2^J$ for some $J$, we can calculate the vector $\bs{f} = [f(1/n),f(2/n),\ldots, f(n/n)]^{\top}$ efficiently via our second ingredient described next.

Any vector $\bs{f} = [f(1/n),f(2/n),\ldots, f(n/n)]^{\top}$, for function $f$ with truncated wavelet basis expansion of order $J$,
can be written as a linear combination of that truncated wavelet basis. In particular, $\bs{f} = W^{\top}\bs{d}$, where $\bs{d} = \left(\alpha_{j_00},\ldots, \alpha_{j_02^{j_0}-1},\beta_{j_00}, \beta_{j_01},\ldots,\beta_{J2^J-1}\right)^{\top}$ is the vector of wavelet coefficients, and the rows of $W$ contain the corresponding wavelet basis functions evaluated at $x_i = i/n$. Specifically, $W$ is an orthogonal matrix with $W_{li} \approx \sqrt{n} \psi_{jk}(i/n)$, or $W_{li} \approx \sqrt{n} \phi_{jk}(i/n)$, for some $l$; the $\sqrt{n}$ factor is due to convention in the literature and software implementation. By orthogonality, $\bs{d} = W\bs{f}$; this transformation from $\bs{f}$ to its wavelet coefficients via multiplication by $W$ is known as the discrete wavelet transform (DWT). The transformation from wavelet coefficients to fitted values, via multiplication by $W^{\top}$ is known as the inverse discrete wavelet transform (IDWT). The DWT and IDWT can be computed in $O(n)$ operations via Mallat's pyramid algorithm~\citep{mallat1989theory}. However, this is only possible for $n=2^J$.

Finally, shrinkage is employed to obtain estimates of the form $\wh{\bm{f}} = W^{\top}\wh{\bm{d}}$; for ease of exposition, we will assume $j_0=0$; i.e., all except the first element of $\bm{d}$ correspond to mother wavelet coefficients. Our methodology and theoretical results do not depend on the choice of $j_0$. 
The wavelet shrinkage estimator is given by
\begin{equation}
\label{eqn:waveShrink}
\wh{\bm{d}} \gets \underset{\bm{d}\in \mathbb{R}^n}{\arg\min \ } \frac{1}{2}\|\bs{y} - W^{\top}\bm{d}\|_2^2 + \lambda \sum_{i=2}^{n}|d_i|,
\end{equation}
for a positive tuning parameter $\lambda$, and given data $\{(i/n, y_i)\in \mathbb{R}^2: i=1,\ldots, n\}$. The $\ell_1$ penalty, $\sum_{i=2}^{n}|d_i| \equiv \|\bm{d}_{-1}\|_1$, shrinks the wavelet coefficients and also induces sparsity; the sparsity is motivated by the desirable \emph{parsimony} property of wavelets: many functions in $L^2([0,1])$ are sparse linear combinations of wavelet bases. The optimization problem \eqref{eqn:waveShrink} can be solved exactly as follows: define $\wt{\bm{d}}  = W\bm{y}$, the DWT of $\bm{y}$. Then, $\wh{d}_1 = \wt{d}_1$ and $\wh{d}_i = sgn(\wt{d}_i)(|\wt{d}_i| - 2\lambda)_+$ ($i=2,\ldots, n$) where $(x)_+ = \max(x,0)$. Thus, for regularly spaced data with $n = 2^J$, wavelet bases provide an efficient nonparametric estimator. In the following subsection, we discuss some existing methods for dealing with irregularly spaced data and present our novel proposal, \name. 

%Numerous extensions of \eqref{eqn:waveShrink} have been proposed, for example, the SURESHRINK procedure of \cite{donoho1995adapting} uses a different tuning parameter for each resolution level $j$, \cite{antoniadis2001regularization} consider various seperable penalties of the form $\sum_{i=2}^np(|d_i|)$, other authors consider block thresholding penalties~\citep{cai1999wavelet,chicken2003block}, for a review of existing variations see \cite{antoniadis1997wavelets}. 

\subsection{A novel interpolation scheme}

The common approach to dealing with irregularly spaced data is to map the observed outcomes $\{ (x_i,y_i)\in [0,1]\times \mathbb{R}:i=1,\ldots,n \}$ to approximate outcomes on the regular grid $\{ (i/n, y^{\prime}_i)\in \mathbb{R}^2:i=1,\ldots,K\}$ for $K = 2^J$ for some integer $J$, via either interpolation or transformation of the data. The novelty of our approach is a reversal of the direction of interpolation, i.e.,  interpolation from fitted values on the regular grid $i/K\ (i=1,\ldots, K)$, to approximated fits on the raw data $x_i\ (i=1,\ldots, n)$. For our proposal, we require an interpolation scheme which can be written as a linear map. In greater detail, for any function $f$ evaluated at a regular grid, $\bm{f} = [f(1/K),\ldots, f(K/K)]^{\top}$ we require an interpolation scheme $\wt{f}(\cdot)$ such that $[\wt{f}(x_1),\ldots, \wt{f}(x_n)]^{\top} = R\bs{f}$ for some interpolation matrix $R\in \mathbb{R}^{n\times K}$. Linear interpolation is a natural choice where
\begin{equation}
\wt{f}(x) = f({i}/{K})\frac{(i+1) - Kx}{(i+1) - i} + f({(i+1)}/{K}) \frac{Kx - i}{(i+1) - i},
\end{equation} for $x\in (i/K, (i+1)/K]$ and $\wt{f}(x) = f(1/K)$ for $x\le 1/K$; and the interpolation matrix is
\begin{align}
\label{eqn:lin-interpol}
	R_{ij} = \left\{ \begin{array}{ll}
			1 & \quad j = 1,  x_i \le 1/K\\
	(j+1) - Kx_i & \quad j = \lfloor Kx_i \rfloor,  x_i \in (1/K, 1] \\
		Kx_i - (j-1) & \quad j = \lceil Kx_i\rceil,  x_i \in (1/K, 1] \\
%		1 & j = K \textbf{ and } x_i > 1\\
	0 & \quad \text{ otherwise }
	\end{array} \right. .
	\end{align} 
Our proposal, \name, solves the following convex optimization problem 
\begin{equation}
\label{eqn:waveMesh}
\wh{\bm{d}} \gets \underset{\bm{d}\in \mathbb{R}^{K}}{\arg\min\ } \frac{1}{2}\|\bs{y} - RW^{\top}\bm{d}\|_2^2 + \lambda \|\bm{d}_{-1}\|_1,
\end{equation}
where $K = 2^{\lceil \log_2 n \rceil }$, $\bm{d}_{-1} = [d_2,\ldots, d_n]^{\top}\in \mathbb{R}^{K-1}$, and $W \in \mathbb{R}^{K\times K}$ is the usual DWT matrix. To evaluate the \name\ estimate at a new point $x\in\mathbb{R}$, one can use $r(x)^{\top}W^{\top}\wh{\bm{d}}$, where $r$ is given by the chosen interpolation scheme. The advantage of \name, over existing methods, is that it can naturally be extended to additive models. Given data $\{(\bs{x}_i,y_i)\in \mathbb{R}^{p+1}: i=1,\ldots, n \}$, let $R_j\in \mathbb{R}^{n\times K}$ be the interpolation matrix corresponding to covariate $j$, i.e., $R_j\bm{f} = [\wt{f}(x_{1j}),\ldots, \wt{f}(x_{nj})]^{\top}$. Then, \name\ can be extended to fitting additive models by the following optimization problem: 
\begin{equation}
\label{eqn:addWaveMesh}
\wh{\bm{d}}_1,\ldots, \wh{\bm{d}}_p \gets \underset{\bm{d}_1,\ldots, \bm{d}_p\in \mathbb{R}^{K}}{\arg\min\ } \frac{1}{2}\Big\|\bs{y} - \sum_{j=1}^p R_jW^{\top}\bm{d}_j\Big\|_2^2 + \lambda\sum_{j=1}^{p} \|\bm{d}_{j,-1}\|_1,
\end{equation}
and $\wh{\bm{f}} = [\wh{f}(\bm{x}_1),\ldots, \wh{f}(\bm{x}_n)]^{\top} =\sum_{j=1}^p \wh{\bm{f}}_j =\sum_{j=1}^p R_jW^{\top}\wh{\bm{d}}_j$. Finally, we can extend additive \name\ to fitting sparse additive models for a potentially large number of covariates. This can be achieved by adding a sparsity inducing penalty for each component $f_j$ as follows:
\begin{equation}
\label{eqn:spaddWaveMesh}
\wh{\bm{d}}_1,\ldots, \wh{\bm{d}}_p \gets \underset{\bm{d}_1,\ldots, \bm{d}_p\in \mathbb{R}^{K}}{\arg\min\ } \frac{1}{2}\Big\|\bs{y} - \sum_{j=1}^p R_jW^{\top}\bm{d}_j\Big\|_2^2 + \sum_{j=1}^{p} \left[ \lambda_1 \|\bm{d}_{j,-1}\|_1 + \lambda_2\|R_jW^{\top}\bm{d}_j\|_2 \right]. 
\end{equation}

\subsection{Algorithm for \name\ and sparse additive \name}

We now present a proximal gradient descent algorithm~\citep{parikh2014proximal} for solving the optimization problem \eqref{eqn:waveMesh}. For convex loss $\ell$ and penalty $P$, the proximal gradient descent algorithm iteratively finds the minimizer of $\left\{ \ell(\bm{d}) + P(\bm{d})\right\}$
%\begin{equation*}
%	\underset{\bm{d}\in \mathbb{R}^K}{\operatorname{minimize}} \ \ell(\bm{d}) + P(\bm{d}),
%\end{equation*}
via the iteration:
\begin{equation*}
\bm{d}^{(l+1)} \gets \underset{\bm{d}\in \mathbb{R}^K }{\arg\min}\ \frac{1}{2} \Big\| \left(\bm{d}^{(l)} - t_l\nabla\ell(\bm{d}^{(l)}) \right)   - \bm{d} \Big\|_2^2 + t_l P(\bm{d}),
\end{equation*} 
for a step-size $t_l>0$. The algorithm is guaranteed to converge as long as $t_l\le L^{-1}$ where $L$ is the Lipschitz constant of $\nabla \ell(\cdot)$. The step-size can be fixed or selected via a line search algorithm. For \eqref{eqn:waveMesh}, we obtain the following iterative scheme:
\begin{equation}
\label{eqn:iter}
\bm{d}^{(l+1)} \gets \underset{\bm{d}\in \mathbb{R}^K }{\arg\min}\ \frac{1}{2} \Big\| \Big\{ (I_K - t_lR^{\top}R)W^{\top}\bm{d}^{(l)} + t_lR^{\top}\bm{y}  \Big\}   - W^{\top}\bm{d} \Big\|_2^2 + t_l \lambda \|\bs{d}^{(l)}_{-1}\|_1.
\end{equation} 
Our algorithm has a number of desirable features which make it computationally efficient. Firstly, \eqref{eqn:iter} is the traditional wavelet problem for regularly spaced data~\eqref{eqn:waveShrink}, with response vector $\bm{r} = \{ (I_K - t_lR^{\top}R)W^{\top}\bm{d}^{(l)} + t_lR^{\top}\bm{y}  \}$. The vector $\bm{r}$ can be efficiently calculated via the sparsity of ${R}$ and Mallat's algorithm for DWT~\citep{mallat1989theory}. Secondly, we can use a fixed step size with $t_l = L^{-1}_{\max}$ where $L_{\max}$ is the maximum eigenvalue of $R^{\top}R$. Again, the maximum eigenvalue can be efficiently computed for sparse matrices, e.g., if $R$ is the linear interpolation matrix then $R^{\top}R$ is tridiagonal, and its eigenvalues can be calculated in $O(K\log K)$ operations. The matrix $R$ for linear interpolation matrix needs to be computed once and requires a sorting of the observations, i.e. $O(n\log n)$. Finally, by taking advantage of Nesterov-style acceleration~\citep{nesterov2007gradient}, the worst-case convergence rate of the algorithm after $k$ steps can be improved from $O(k^{-1})$ to $O(k^{-2})$.

The procedure \eqref{eqn:iter} can also be used to solve the additive~\eqref{eqn:addWaveMesh} and sparse additive~\eqref{eqn:spaddWaveMesh} extensions via a block coordinate descent algorithm. Specifically, given a set of estimates $\bs{d}_j$ ($j=1,\ldots,p$) we can fix all but one of the vectors $\bs{d}_j$ and optimize over the non-fixed vector, by solving
\begin{equation}
\label{eqn:proxHD}
\underset{\bs{d}\in \mathbb{R}^K}{\operatorname{minimize }}\ \frac{1}{2}\|\bm{r}_j - R_jW^{\top}\bs{d}\|_2^2 + \lambda_1\|\bs{d}_{-1}\|_1 + \lambda_2\|R_jW^{\top}\bm{d}\|_2,
\end{equation} 
for some vector $\bm{r}_j\in \mathbb{R}^n$. For additive \name~($\lambda_2 = 0$), this reduces to the univariate problem which can be solved via the algorithm \eqref{eqn:iter}. For sparse additive \name~($\lambda_2\not=0$), the problem can be solved by solving \eqref{eqn:proxHD} with $\lambda_2=0$ following by a soft-scaling operation~\citep[Lemma 7.1]{petersen2016fused}. We detail our algorithm for sparse additive \name\ in the supplementary material.  

\subsection{Some extensions and variations}
In this subsection, we discuss some variations and extensions of \name, namely (1) using a conservative order for the wavelet basis expansion, (2) extending \name\ for more general loss functions and, (3) using a weighted $\ell_1$ penalty for shrinkage of wavelet coefficients.  

While in \eqref{eqn:waveMesh} we set $K = 2^{\lceil \log_2  n \rceil}$, we could, instead, set $K$ to be any power of 2. Since the main computational step in our algorithm is the DWT and IDWT which requires $O(K)$ operations, a smaller value of $K$ can greatly reduce the computation time. Furthermore, using a smaller $K$ can lead to superior predictive performance in some settings; this is formalized in our theoretical results of Section~\ref{sec:theory} and observed in the simulation studies of Section~\ref{sec:simulations}. In the supplementary material we present additional simulation studies comparing the prediction performance and computation time of \name\ for various values of $K$.

Secondly, \name\ can be extended to other loss functions appropriate for various data types. For example, we can extend our methodology to the setting of binary classification via a logistic loss function. Let $y_i\in \{-1,1\}\ (i=1,\ldots,n)$ be the observed response. For the univariate case, we get
\begin{equation}
\label{eqn:waveMeshLogit}
\wh{\bm{d}} \gets \underset{\bm{d}\in \mathbb{R}^{K}}{\arg\min\ } \frac{1}{2}\sum_{i=1}^{n} \log\left( 1 +\exp\left[-y_i(RW^{\top}\bm{d})_i\right] \right) + \lambda \|\bm{d}_{-1}\|_1.
\end{equation}
Like the least squares loss, \eqref{eqn:waveMeshLogit} naturally extends to (sparse) additive models. The problem can be efficiently solved via a proximal gradient descent algorithm described in the supplementary material. 

Finally, we consider a variation of our $\ell_1$ penalty motivated by the \texttt{SURESHRINK} procedure of \cite{donoho1995adapting}. For a vector $\bs{d}\in \mathbb{R}^K$ of discrete father and mother wavelet coefficients, denote by $\bs{d}_{[j]}$ the discrete mother wavelet coefficients at resolution level $j$. For this particular variation, we require that the minimum resolution level $j_0>1$. We then propose to solve
\begin{equation}
\label{eqn:adapWaveMesh}
\wh{\bm{d}} \gets \underset{\bm{d}\in \mathbb{R}^{K}}{\arg\min\ } \frac{1}{2}\|\bs{y} - RW^{\top}\bm{d}\|_2^2 + \lambda \sum_{j=j_0}^{\log_2 K}\sqrt{2\log(j)}\|\bs{d}_{[j]}\|_1.
\end{equation}
In the supplementary material we show that the above estimator outperforms the usual \name\ estimator \eqref{eqn:waveMesh} in terms of prediction error.

% !TeX root = ../Manuscript.tex

\section{Theoretical results}
\label{sec:theory}

In this section, we study finite sample properties of our univariate estimator \eqref{eqn:waveMesh}, and sparse additive estimator~\eqref{eqn:spaddWaveMesh}. We begin with a quick introduction to Besov spaces and their connection to wavelet bases. We establish minimax convergence rates (up to a $\log n$ factor) for our univariate proposal.  We note that our estimator \eqref{eqn:waveMesh} can be seen as a lasso estimator~\citep{tibshirani1996regression} with design matrix $RW^{\top}$; this allows us to use well-known results for the lasso estimator to easily establish minimax rates which we present below. Additionally, the lasso formulation allows us to establish sufficient conditions for the  uniqueness of our estimator. Specifically, fitted values $\wh{\bs{f}} = RW^{\top} \wh{d}$ are unique whereas uniqueness of $\wh{d}$ depends on the matrix $RW^{\top}$. In the interest of brevity, we omit derivation of sufficient conditions for uniqueness of $\wh{d}$ and refer the interested reader to~\cite{tibshirani2013lasso}. Finally, we also establish rates for the sparse additive \name\ proposal for a specific penalty. 

Besov spaces on the unit interval, $B^s_{q_1,q_2}$, are function spaces with specific degrees of smoothness in their derivative: for the Besov norm $\|\cdot\|_{B^s_{q_1,q_2}}$, 
%\begin{equation}
%\label{eqn:besovSpace}
$B_{q_1,q_2}^s = \{ g\in L^2([0,1]): \|g\|_{B^s_{q_1,q_2}} < C \}.$
%\end{equation}
The constants $(s,q_1,q_2)$ are the parameters of Besov spaces; for a function $g\in L^2([0,1])$ with the wavelet bases expansion \eqref{eqn:waveletExpansion}, the Besov norm is defined as 
\begin{equation}
\label{eqn:besovNorm}
\|g\|_{B^s_{q_1,q_2}} = \|\bm{\alpha}_{j_0}\|_{q_1} + \bigg[ \sum_{j=j_0}^{\infty}\bigg\{2^{j(s+1/2-1/q_1)}\|\bs{\beta}_j\|_{q_1}\bigg\}^{q_2} \bigg]^{1/q_2},
\end{equation}
where $\bm{\alpha}_{j_0}\in \mathbb{R}^{2^{j_0}}$ is the vector of father wavelet coefficients with minimum resolution level $j_0$ and $\bm{\beta}_{j}\in \mathbb{R}^{2^j}$ is the vector of mother wavelet coefficients at resolution level $j$. For completeness, we also define $\|g\|_{B^s_{q_1,\infty}} = \|\bm{\alpha}_{j_0}\|_{q_1} + \sup_{j\ge j_0}\left\{2^{j(s+1/2-1/q_1)}\|\bs{\beta}_j\|_{q_1}\right\} $. We consider Besov spaces because they generalize well-known classes such as the Sobolev ($B^s_{2,2},\ s = 1,2,\ldots$), and H\"{o}lder ($B_{\infty, \infty}^s,\, s>0$) spaces and the class of bounded total variation functions (sandwiched between $B_{1,1}^1$ and $B_{1,\infty}^1$). Our first result below establishes near minimax convergence rates for the prediction error of our estimator. An attractive feature of our estimator is that it achieves this rate without any information about the parameters $(s,\ q_1,\ q_2)$. We recover the usual wavelet rates of \cite{donoho1995noising} under the special case when $x_i = i/n$ and $R = I_n$. Additionally, the theorem justifies the use of $K<n$ basis functions: if the true function is sufficiently smooth, we recover the usual rates with an additional $\log K$ factor instead of $\log n$.
\begin{theorem}
	\label{thm:MainThm}
	Suppose $y_i = f^0(x_i) + \e_i\ (i=1,\ldots, n)$ for mean zero, sub-Gaussian noise $\e_i$. Define the estimator $\wh{\bm{f}} = RW^{\top}\wh{\bm{d}} = [\wh{f}(x_1),\ldots, \wh{f}(x_n)]^T$ for linear interpolation matrix $R$~\eqref{eqn:lin-interpol} where 
	\begin{equation*}
	\wh{\bm{d}} \gets \underset{\bm{d}\in \mathbb{R}^K}{\arg\min }\ \frac{1}{2}\|\bs{y} - RW^{\top}\bm{d}\|_2^2 + \lambda\|\bm{d}_{-1}\|_1,
	\end{equation*}
	for the usual DWT transform matrix $W\in \mathbb{R}^{K\times K}$ associated with some orthogonal wavelet family. Further, define $\bm{f}^0 = [f^0(x_1),\ldots, f^0(x_n)]^{\top}$ and $\wt{\bm{f}}^0 = [f^0(1/K),\ldots, f^0(K/K)]^{\top}$. Assume that $f^0\in B^s_{q_1,q_2}$ and the mother wavelet $\psi$, has $r$ null moments and $r$ continuous derivatives where $r>\max\{1,s\}$. Suppose $\lambda\ge c_1\sqrt{t^2 + 2\log K}$ for some $t>0$. Then, for sufficiently large $K$ (specifically $K\ge c_1n^{1/(2s+1)}$ for some constant $c_1$), with probability at least $1-2\exp(-t^2/2)$, we have 
	\begin{equation*}
	\frac{1}{n}\left\| \bm{f}^0 - \wh{\bm{f}} \ \right\|_2^2 \le C\left(\frac{\log K}{n}\right)^{\frac{2s}{2s+1}} + \frac{2}{n}\|\bm{f}^0 - R\wt{\bm{f}}^0\|_2^2,
	\end{equation*}
	where the constant $c_1$ depends on $R$ and the distribution of $\e_i$, and the constant $C$ depends on $R$.
\end{theorem} 

The above theorem includes an approximation error term $\|\bm{f}^0 - R\wt{\bm{f}}^0\|_2^2$ which depends on the type of interpolation matrix $R$. For example, for linear interpolation of a twice continuously differentiable function, the approximation error scales as $O(K^{-2})$. Thus, for a sufficiently large $K$ (particularly $K=n$), the approximation error will disappear. In fact, as long as the approximation error is of the order $(\log K/n)^{2s/(2s+1)}$, we obtain the usual near-minimax rate.

For the sparse additive model, we consider a different model motivated by the Besov norm \eqref{eqn:besovNorm}. Our next theorem provides convergence rates for the estimated function $\wh{\bm{f}} = \sum_{j=1}^p\wh{\bm{f}}_j = \sum_{j=1}^{p}R_jW^{\top}\wh{\bm{d}}_j$, where
\begin{equation}
\label{eqn:spaddWaveMesh2}
\wh{\bm{d}}_1,\ldots, \wh{\bm{d}}_p \gets \underset{\bm{d}_1,\ldots, \bm{d}_p\in \mathbb{R}^{K}}{\arg\min\ } \frac{1}{2}\Big\|\bs{y} - \sum_{j=1}^p R_jW^{\top}\bm{d}_j\Big\|_2^2 + \sum_{j=1}^{p} \left[ \lambda_1 P_{s}(\bm{d}_j) + \lambda_2\|R_jW^{\top}\bm{d}_j\|_2 \right],
\end{equation}
and the penalty $P_{s}$ is the discrete version of the Besov norm for $B_{1,1}^s$. Specifically, for $\bm{d}$ as a vector of father coefficients, $\alpha_{j_0k}\ (k=0,\ldots,2^{j_0}-1)$, and mother wavelet coefficients $\beta_{jk}\ (j=j_0,\ldots, J; k= 0,\ldots,2^{j}-1)$ the penalty is 
\begin{equation}
\label{eqn:newPen}
P_s(\bm{d}) = \sum_{k=0}^{2^{j_0}-1}|\alpha_{j_0k}| + \sum_{j=j_0}^J\Big( 2^{j(s-1/2)}\sum_{k=0}^{2^j-1}|\beta_{jk}| \Big).
\end{equation} 
Before presenting our next result, we state and discuss the so called \emph{compatibility} condition. This condition is common in the high-dimensional literature~\citep{geer2009conditions} and crucial for proving minimax rates for sparse additive models. Briefly, our proof requires the semi-norms $\sum_{j\in S} \|f_j\|_2$ and $\|\sum_{j=1}^{p}f_j\|_2$ to be  somehow `compatible', for an index set $S\subseteq \{1,\ldots, p\}$. In the low-dimensional/non-sparse case, i.e., $S = \{1,\ldots,p\}$, the semi-norms are compatible by the inequality $\sum_{j\in S}\|f_j\|_2\le \sqrt{|S|}\|\sum_{j=1}^{p}f_j\|_2$. The compatibility condition ensures such an inequality holds for proper subsets $S$. Furthermore, the compatibility condition can be relaxed at the cost of proving a slower rate; this is similar to the lasso \emph{slow rate}~\citep{dalalyan2017prediction}.

\begin{definition}
	The compatibility condition is said to hold for an index set $S\subset\{1,2,\ldots,p\}$, with compatibility constant $\vartheta(S)>0$, if for all $\gamma>0$ and any set of discrete wavelet coefficients vector $(\bm{d}_1,\ldots, \bm{d}_p)$, that satisfy $\sum_{j\in S^c}n^{-1}\|R_jW^{\top}\bm{d}_j\|_2 + \gamma\sum_{j=1}^{p}P_s(\bm{d}_j) \le 3\sum_{j\in S}\|R_jW^{\top}\bm{d}_j\|$, it holds that
%	\begin{equation}
%	\label{eqn:compCond}
$	\sum_{j\in S} \|R_jW^{\top}\bm{d}_j\|_2 \le \sqrt{|S|}\Big\|\sum_{j=1}^pR_jW^{\top}\bm{d}_j \Big\|_2/\vartheta(S).$
%	\end{equation}
\end{definition}

\begin{theorem}
	Assume the model $y_i = f^0(\bs{x}_i) + \e_i$ ($i=1,\ldots, n$) with mean zero, sub-Gaussian $\e_i$. Let $\wh{f} = \sum_{j=1}^{p}\wh{\bm{f}}_j$ be as defined in \eqref{eqn:spaddWaveMesh2}, and let $\bm{f}^* = \sum_{j\in S^*} \bm{f}_j^* = \sum_{j\in S^*} R_jW^{\top}\bm{d}_j^*$ be an arbitrary sparse additive function with $S^*\subset \{1,2,\ldots, p\}$. Let $\rho = \kappa \max \{ n^{-2s/(2s+1)}, (\log p/n)^{1/2} \}$ for a constant $\kappa$ that depends on the distribution of $\e_i$ and $s$. Suppose $\lambda\ge 4\rho$. Then, with probability at least $1 -2 \exp(-c_1n\rho^2) - c_2\exp(-c_3n\rho^2)$, we have
	\begin{equation*}
	n^{-1}\left\|\bm{f}^0 - \wh{\bm{f}}\ \right\|_2^2 \le C_1\max\Big\{ |S^*|n^{-\frac{s}{2s+1}} , |S^*|\Big( \frac{\log p}{n}\Big)^{1/2} \Big\} + n^{-1}\left\|\bm{f}^0 - \bm{f}^*\right\|_2^2,
	\end{equation*}
	where constants $c_1, c_2$ depend on the distribution of $\e_i$ and $s$, and $C_1$ depends on $\kappa$ and $|S^*|^{-1}\sum_{j\in S^*}P_{s}(\bm{d}_j^*)$.	Furthermore, if the compatibility condition holds for $S^*$ with constant $\vartheta(S^*)$ we have 
	\begin{equation*}
	n^{-1}\left\|\bm{f}^0 - \wh{\bm{f}}\ \right\|_2^2 \le C_2\max\Big\{ |S^*|n^{-\frac{2s}{2s+1}} , |S^*| \frac{\log p}{n} \Big\} + 4n^{-1}\left\|\bm{f}^0 - \bm{f}^*\right\|_2^2,	
	\end{equation*}
	where the constant $C_2$ depends on $\vartheta(S^*)$ and $|S^*|^{-1}\sum_{j\in S^*}P_{s}(\bm{d}_j^*)$.
\end{theorem}

% !TeX root = ../Manuscript.tex

\section{Numerical experiments}
\label{sec:simulations}

\subsection{Experiments for univariate regression}
\label{sec:simulationUni}
We begin with a simulation to compare the performance of univariate \name\ to the traditional interpolation method of \cite{kovac2000extending}, isometric wavelet method of \cite{sardy1999wavelet}---which treats the data as if it were regularly spaced---and adaptive lifting method of \cite{nunes2006adaptive}. The former two methods are implemented in the \texttt{R} package \texttt{wavethres}~\citep{nason2016wavethresh} and the latter is implemented in the \texttt{adlift} package~\citep{nunes2017adlift}.

We generate the data as $y_i = f^0(x_i) + \e_i$ ($i=1,\ldots, n$) for different choices of function $f^0$ and $n$. The errors are distributed as $\e_i\sim \mathcal{N}(0,\sigma^2)$ with $\sigma^2$ chosen such that $\operatorname{SNR} = 5$, where $\operatorname{SNR} = \operatorname{var}(\bs{f}^0)/\sigma^2$. We consider two different choices of the covariate, $x_i\sim \mathcal{U}[0,1]$ and $x_i\sim \mathcal{N}(0, 1)$ scaled to lie in $[0,1]$. We consider 6 different choices for the function $f^0$: 1. polynomial, 2. sine, 3. piecewise polynomial, 4. heavy sine, 5. bumps and, 6. doppler. These functions are shown in Figure~1 of the supplementary material. We apply our proposal, {\name}, the interpolation proposal of \cite{kovac2000extending} and isometric wavelet proposal of \cite{sardy1999wavelet}, for a sequence of 50 $\lambda$ values linear on the log scale and select the $\lambda$ value that minimizes the mean square error, $\operatorname{MSE} = n^{-1}\big\|\bm{f}^0 - \wh{\bm{f}} \big\|_2^2$. For adaptive lifting, the \texttt{R} implementation automatically selects a tuning parameter. We implement \name\ using the linear interpolation matrix \eqref{eqn:lin-interpol}. We also implement \name\ using a small grid, i.e., we fit \eqref{eqn:waveMesh} with $K = 2^{5}$ and $2^6$. The \texttt{R} implementation of isometric wavelets requires sample sizes to be a power of two; if not, we pad the response vector with zeros. 

We also analyze the motorcycle data studied by \cite{silverman1985some} consisting of 133 head acceleration measurements in a simulated motorcycle accident taken at 94 unequally spaced time points. To avoid the issue of repeated measurements, we average acceleration measurements at the same time leading to a sample size of $n = 94$. Selection of tuning parameter for \name\ is done via 5-fold cross validation. For interpolation~\citep{sardy1999wavelet} and isometric~\citep{kovac2000extending} wavelet proposals, we use the \emph{universal thresholding} rule for tuning parameter selection~\citep{donoho1994ideal}; this rule leads to near minimax convergence rates like that of Theorem~\ref{thm:MainThm}. 

%\begin{figure}
%	\includegraphics[scale = 0.5]{plots/TrueFuncs2a.pdf}
%	\caption{Plots of functions $f^0$ for the simulation study. Functions in green are the most smooth and well-behaved followed by functions with moderate smoothness in orange. Finally, functions in red are highly irregular functions, e.g., functions with unbounded total variation.}
%	\label{fig:truef}
%\end{figure}

Table~\ref{tab:unif} shows the ratio of $\operatorname{MSE}$ between our proposal with $K = 2^{\lceil \log_2 n\rceil}$ and other proposals for uniformly distributed $x_i$. We observe that our proposal has the smallest MSE for all functions except the Bumps function. Even for the Bumps function, \name\ exhibits superior prediction performance over other methods for $n = 512$. We also observe that \name\ with smaller values of $K$ often outperforms the full \name~($K = 2^{\lceil \log_2 n\rceil}$) method in terms of MSE. 
Results for normally distributed $x_i$ are given in the supplementary material. In that case, we again observe that \name\ outperforms existing methods for a number of simulation scenarios, except for a few cases with polynomial and bumps functions. Results for sample sizes that are not powers of two were similar to the results provided here. In the interest of brevity, these results are presented in the supplementary material.

\begin{table}
	\caption{Results for $x_i\sim \mathcal{U}[0,1]$ averaged over 100 replicates; the ratio $\operatorname{MSE}/\operatorname{MSE}_{FG}$ is shown along with $100\times$ the standard error, where $\operatorname{MSE}_{FG}$ is the $\operatorname{MSE}$ of \name\ with $K = 2^{\lceil \log_2 n\rceil}$. Boldface values represent the method with the smallest MSE within each row of the table.}
	\label{tab:unif}
	\begin{tabular}{llccccc}
		\hline
		& & \name & \name & Interpolation & Isometric & \multicolumn{1}{c}{Adaptive Lifting} \\ 
		& & $K = 2^5$ & $K = 2^6$ &  &  & \\
		\hline
		Polynomial & n = 64  & 1.19 (5.51) & \textbf{1.00} (0.00) & 1.24 (4.11) & 1.78 (7.56) & 4.28 (29.86) \\
		& n = 128  & 0.92 (5.57) & \textbf{0.77} (3.07) & 1.12 (6.00) & 1.33 (7.18) & 3.57 (31.27) \\
		& n = 256  & 1.00 (6.20) & \textbf{0.85} (3.15) & 1.61 (9.04) & 1.50 (7.67) & 4.29 (31.29) \\
		& n = 512  & 0.78 (3.18) & \textbf{0.72} (2.58) & 1.76 (6.11) & 1.13 (2.64) & 3.61 (26.47) \\
		Sine & n = 64  & \textbf{0.97} (3.14) & 1.00 (0.00) & 1.47 (5.81) & 1.59 (6.72) & 3.62 (33.65) \\
		& n = 128  & \textbf{0.76} (3.18) & 0.76 (1.96) & 1.29 (6.08) & 1.46 (5.24) & 2.98 (19.78) \\
		& n = 256  & \textbf{0.66} (2.50) & 0.70 (2.22) & 1.93 (9.49) & 1.34 (4.23) & 3.41 (18.80) \\
		& n = 512  & 0.57 (2.34) & \textbf{0.56} (2.22) & 2.13 (7.78) & 1.24 (3.66) & 3.63 (28.42) \\
		\hline
		Piecewise & n = 64  & \textbf{0.85} (1.97) & 1.00 (0.00) & 1.18 (3.12) & 1.31 (3.62) & 1.63 (9.07) \\
		Polynomial & n = 128  & \textbf{0.77} (2.00) & 0.82 (1.52) & 1.26 (2.75) & 1.22 (2.61) & 1.40 (7.36) \\
		& n = 256  & 0.82 (1.92) & \textbf{0.79} (1.59) & 1.42 (3.18) & 1.14 (2.11) & 1.15 (6.04) \\
		& n = 512  & 1.01 (2.43) & \textbf{0.86} (1.70) & 1.71 (3.56) & 1.15 (1.99) & 1.25 (7.24) \\
		Heavy Sine & n = 64  & \textbf{0.84} (2.44) & 1.00 (0.00) & 1.12 (3.04) & 1.41 (3.17) & 1.70 (8.35) \\
		& n = 128  & \textbf{0.75} (2.66) & 0.82 (1.16) & 1.17 (3.32) & 1.50 (4.75) & 1.56 (8.26) \\
		& n = 256  & \textbf{0.66} (1.64) & 0.72 (1.14) & 1.37 (2.98) & 1.33 (2.58) & 1.53 (6.74) \\
		& n = 512  & \textbf{0.58} (1.59) & 0.60 (1.18) & 1.58 (3.05) & 1.29 (1.60) & 1.50 (9.21) \\
		\hline
		Bumps & n = 64  & 2.11 (2.30) & 1.00 (0.00) & 1.70 (1.75) & \textbf{0.72} (1.34) & 1.07 (5.12) \\
		& n = 128  & 2.86 (2.77) & 2.11 (1.62) & 1.40 (1.59) & \textbf{0.63} (0.83) & 0.85 (2.43) \\
		& n = 256  & 4.81 (6.82) & 3.47 (4.39) & 1.43 (1.89) & \textbf{0.88} (0.99) & 0.97 (2.00) \\
		& n = 512  & 7.45 (9.13) & 5.69 (6.77) & 1.32 (1.35) & \textbf{1.19} (1.03) & 1.23 (2.34) \\
		Doppler & n = 64  & \textbf{0.98} (1.69) & 1.00 (0.00) & 1.15 (3.45) & 1.33 (3.20) & 1.30 (3.65) \\
		& n = 128  & 1.24 (2.02) & \textbf{0.89} (1.04) & 1.07 (2.13) & 1.44 (2.57) & 1.18 (3.22) \\
		& n = 256  & 1.71 (3.92) & \textbf{0.94} (1.38) & 1.20 (2.11) & 1.29 (1.99) & 1.30 (3.44) \\
		& n = 512  & 2.58 (4.85) & 1.26 (2.01) & 1.21 (1.48) & \textbf{1.10} (1.31) & 1.23 (3.36) \\
		\hline 
	\end{tabular}
\end{table}

\begin{figure}
	\centering
	\includegraphics[width=0.45\textwidth]{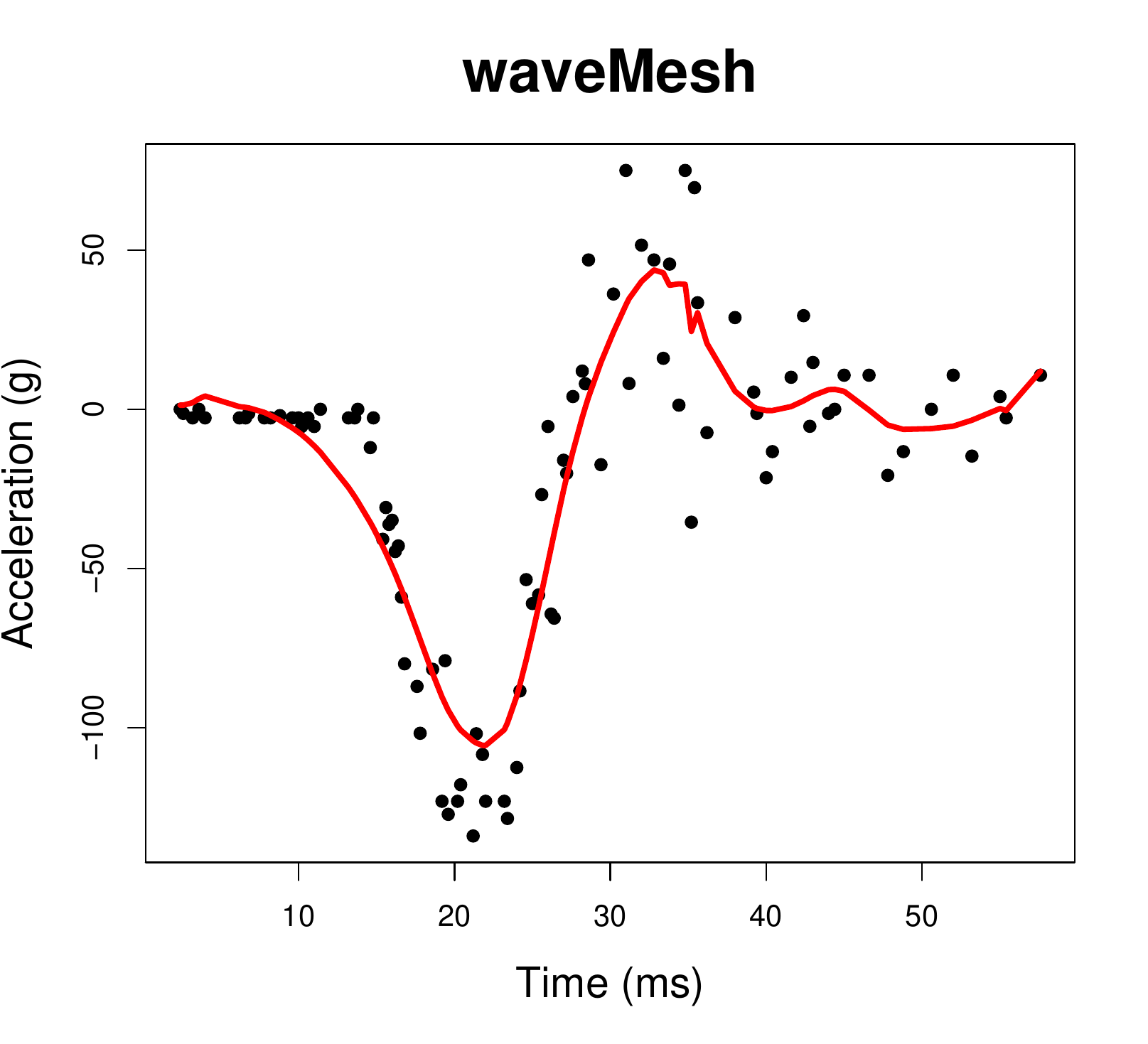}
	\includegraphics[width=0.45\textwidth]{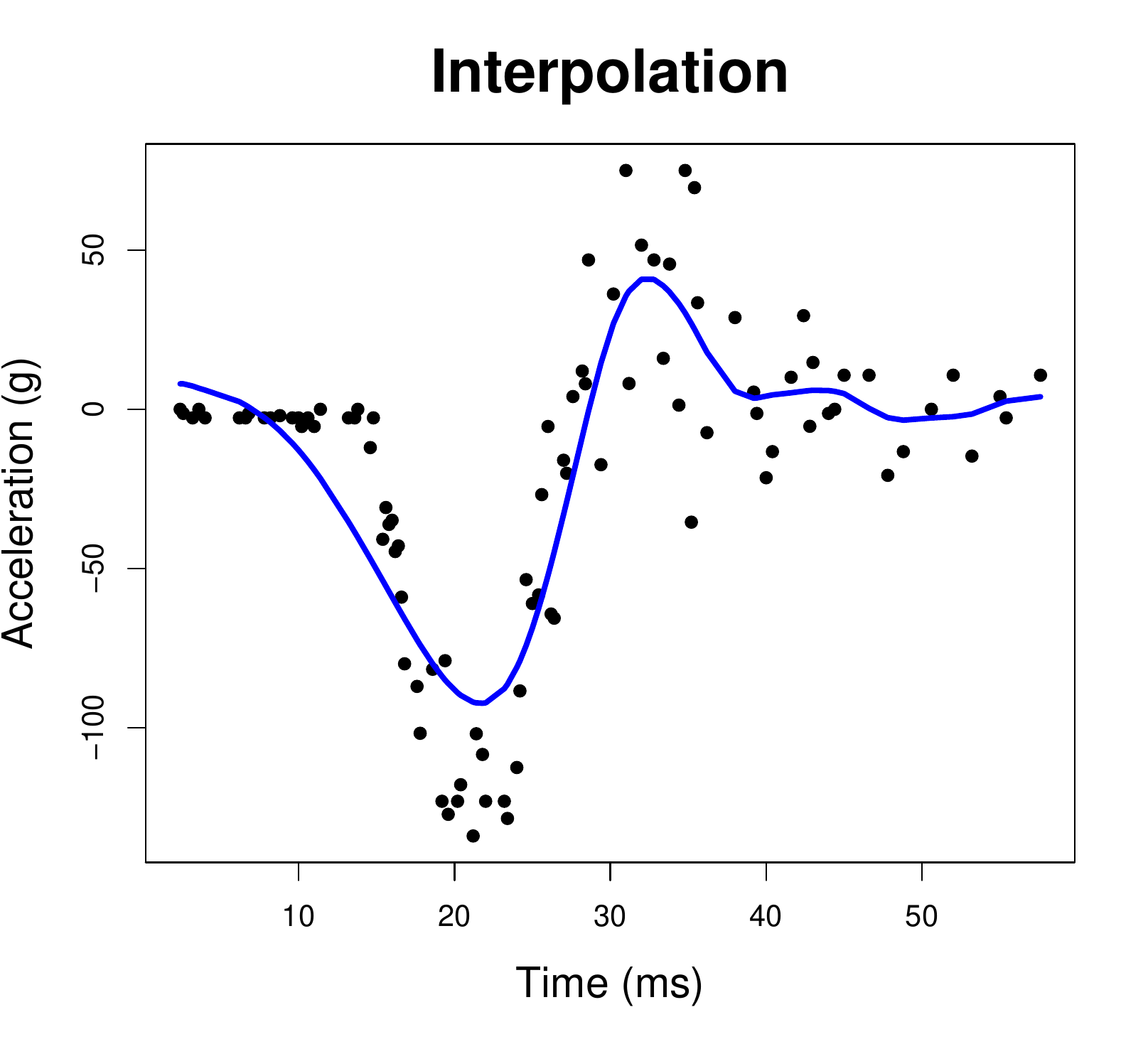}
	\includegraphics[width=0.45\textwidth]{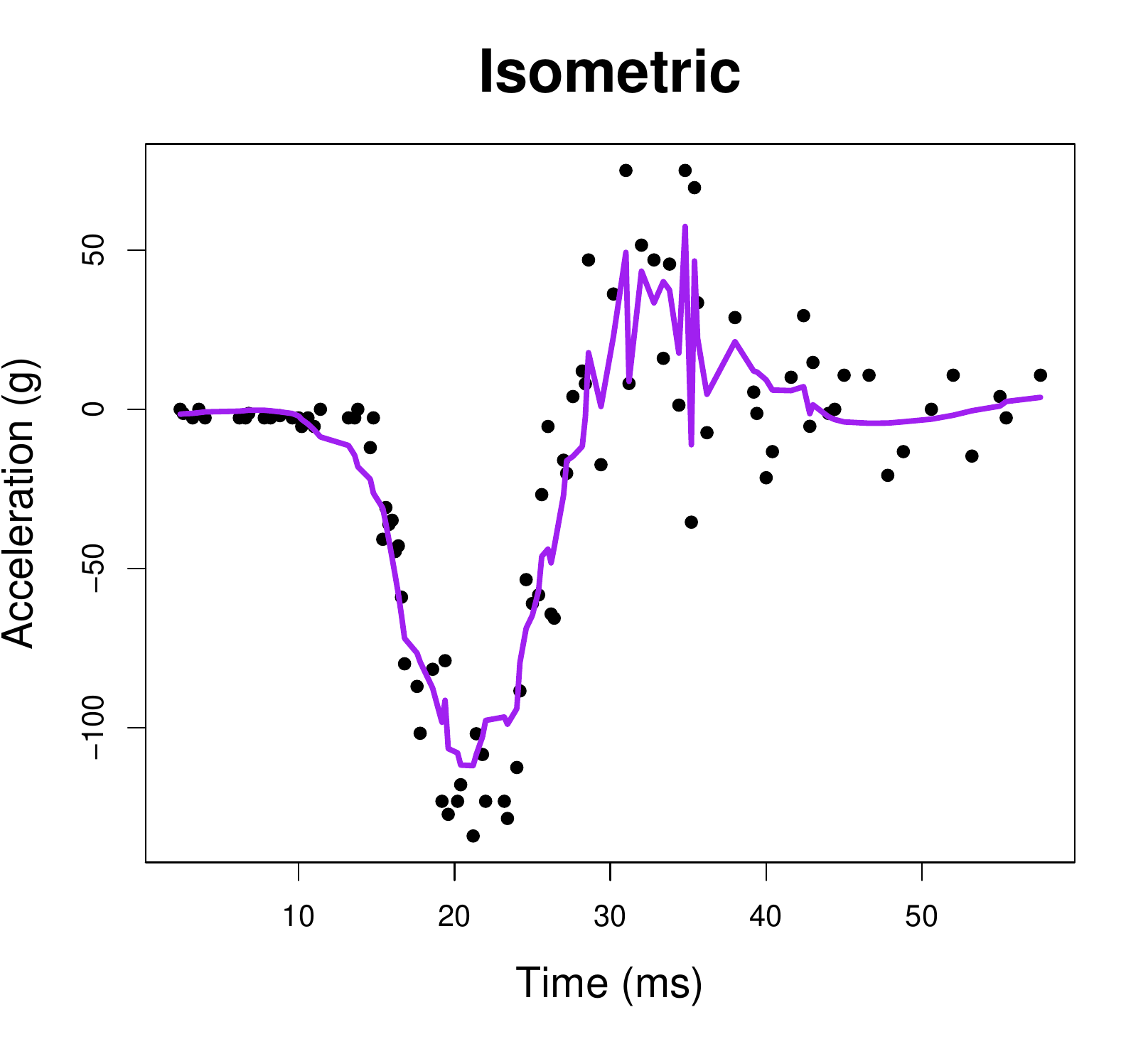}
	\includegraphics[width=0.45\textwidth]{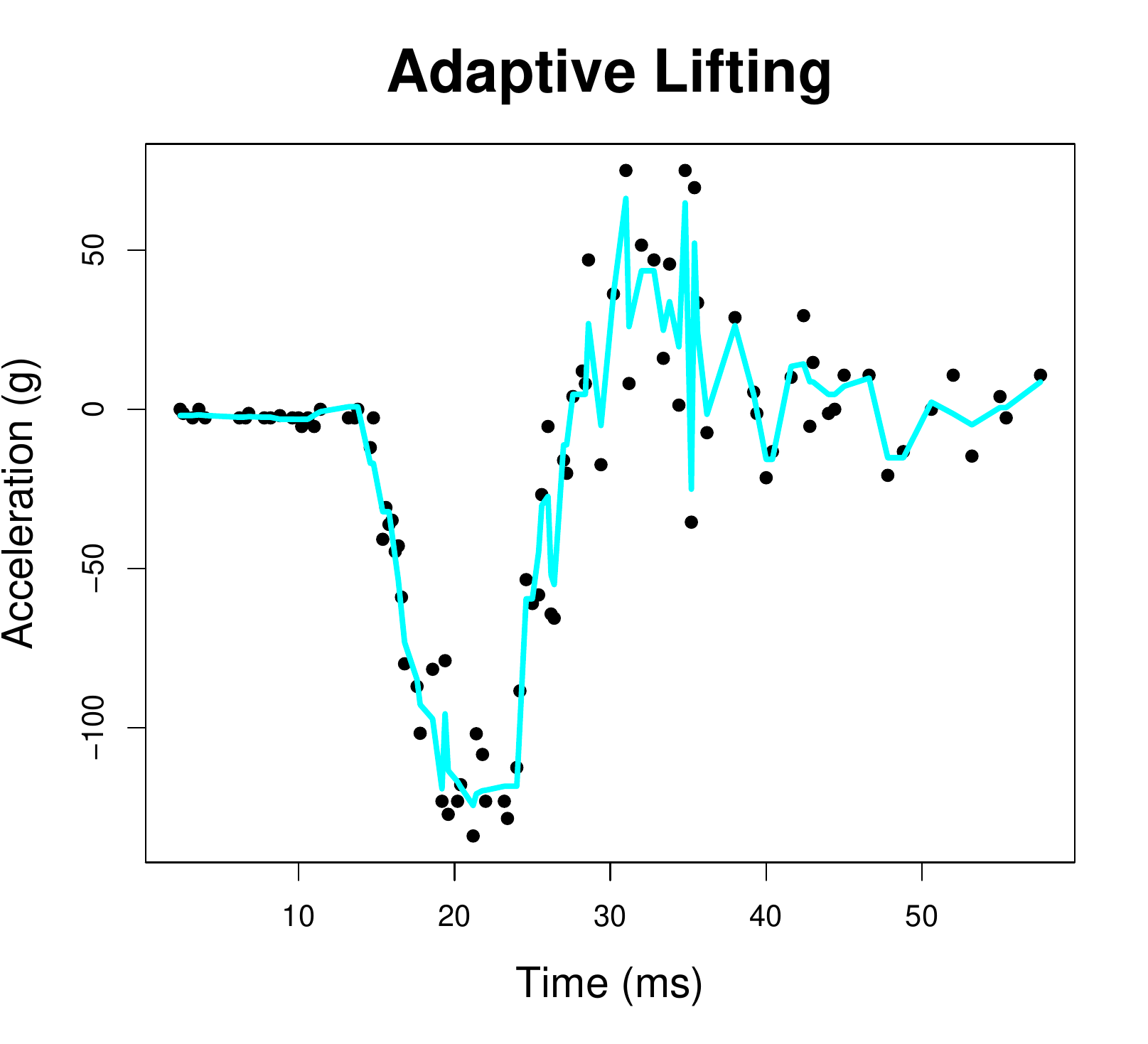}
	\caption{Fitted functions to the motorcycle accident dataset for each of the 4 methods.}
	\label{fig:dataUni}
\end{figure}

In Figure~\ref{fig:dataUni}, we plot the motorcycle data and fitted functions for each method. Here, \name\ reasonably models the data via a smooth function; the interpolation method has a similar but slightly more biased result around 10 to 25 ms. Adaptive lifting and isometric wavelets lead to highly variable estimates.

\subsection{Experiments for multivariate additive regression}

We proceed with a simulation study to illustrate the performance of additive \name\ compared to the proposal of \cite{sardy2004amlet}, AMlet. We use the author-provided \texttt{R} implementation for the AMlet proposal; due to a lack of \texttt{R} packages for other proposals, we defer the comparison to future work. We consider the following simulation setting: we generate data with $y_i  = f_1(x_{i1}) + f_2(x_{i2}) + f_{3}(x_{i3}) + f_4(x_{i4}) + \e_i$ ($i=1,\ldots, 2^{10}$), where $\e_i\sim \mathcal{N}(0, \sigma^2)$, $x_i\sim \mathcal{U}[0,1]$, and $\sigma^2$ such that $\operatorname{SNR}=10$. The four functions $f_1,\ldots, f_4$ are the polynomial, sine, piecewise polynomial and heavy sine functions presented in Figure~1 of the supplementary material. We consider sample sizes $n = 64, 100, 256, 500, 512$ and results were averaged over 100 data sets. 
%The existing \texttt{R} implementation of AMlet requires sample size to be a power of 2; 
For sample sizes not a power of 2, the response vector was padded with zeros for the \texttt{R} implementation of AMlet. 
The universal threshold rule was used for AMlet as detailed in \cite{sardy2004amlet}; 5-fold cross validation was used for additive \name\ for selection of $\lambda$.

For a real world data analysis, we consider the Boston housing data analyzed by \citet{ravikumar2009sparse}. The goal is to predict the median value of homes based on 10 predictors. The data consists of $n = 506$ observations; we use $256$ observations for training and calculate the test error on the rest. Tuning parameters are selected in the same way as the simulation study above. 

Table~\ref{tab:additive} shows the MSE of both proposals for various choices of $n$ for the simulation study. The results clearly indicate that additive \name\ offers substantial improvement over AMlet, especially for smaller values of $n$.  We observe similar results for the Boston housing data: the average test error is 21.2 for \name\ (standard error 0.34) and 25.1 for AMlet (standard error 0.42). These results support our theoretical analysis and underscore the advantages of \name\ in sparse high-dimensional additive models. 

\begin{table}[b]
	\centering
		\caption{MSE and standard error of \name\ and AMlet averaged over 100 data sets.}
		\label{tab:additive}
		\begin{tabular}{lcccccc}
			\hline
			&  $n=64$ & $n = 100$ & $n=128$ & $n=256$&  $n = 500$  & $n=512$ \\
			\hline
			\name & \textbf{10.76} (0.31)& \textbf{11.35} (0.33) & \textbf{8.82} (0.24) & \textbf{5.45} (0.11) &  \textbf{4.34} (0.08) & \textbf{4.08} (0.07)\\
			AMlet & 100.48 (1.83) &   34.58 (1.05) &  45.49  (1.09) & 19.57 (0.33) & 10.67 (0.12)  & 8.90 (0.11)\\
			\hline
		\end{tabular}
\end{table}

%\input{sections/realdata}
% !TeX root = ../Manuscript.tex

\section{Conclusion}
\label{sec:conclusion}

In this paper, we introduced \name, a novel method for non-parametric regression using wavelets. Unlike traditional methods, \name\ does not require that covariates are uniformly spaced on the unit interval, nor does it require that  the sample size is a power of 2. We achieve this using a novel interpolation approach for wavelets. The main appeal of our proposal is that it naturally extends to multivariate additive models for a potentially large number of covariates.

To compute the estimator, we proposed an efficient proximal gradient descent algorithm, which leverages existing techniques for fast computation of the DWT. We established minimax convergence rates for our univariate proposal over a large class of Besov spaces. For a particular Besov space, we also established minimax convergence rates for our (sparse) additive framework. The \texttt{R} package \texttt{waveMesh}, which implements our methodology, will soon be publicly available on GitHub. 

\newpage

\subsubsection*{Acknowledgments}
We thank three anonymous referees for insightful comments that substantially improved the manuscript. We thank Professor Sylvain Sardy for providing software. This work was partially supported by National Institutes of Health grants to A.S. and N.S., and National Science Foundation grants to A.S.

\bibliographystyle{plainnat}
\bibliography{bibfileah/refnew}

\end{document}

	
	\maketitle
	
	% !TeX root = ../MANUSCRIPT.tex

\section{Details of Algorithms}
\label{sec:algs}

Here we give an algorithm for our additive and sparse-additive framework as well as an algorithm for the extension of our proposal to classification. We use a block-wise coordinate descent algorithm for solving the additive and sparse additive proposal. This algorithm cyclically iterates through features, and for each feature applies the univariate solution detailed in the main manuscript. The exact details are given in Algorithm 1 below.

\begin{algorithm}
	\caption{Block coordinate descent for the additive and sparse additive framework}
	\label{alg:additive}
	\vspace*{-12pt}
	\begin{tabbing}
		\enspace Initialize ${\bm{d}}_j \gets 0$ for $j=1,\ldots,p$\\
		\enspace While $l\le max\_iter$ \textbf{ and } not converged\\
		
		\qquad For $ j = 1,\, \ldots,\, p$\\
		\qquad \qquad Set 
		$ {\bm{r}}_{-j} \gets {\bm{y}} - \sum_{j^{'} \neq j} R_{j'}W^{\top}\bm{d}_{j'} $\\
		\qquad \qquad Update $\bm{d}_j \gets \underset{\bm{d} \in \mathbb{R}^{\K}}{\arg\min} \frac{1}{2} \left\| {r}_{-j} - R_jW^{\top} \bm{d} \right\|_2^2 + \lambda_1\|\bm{d}_{-1}\|_1 + \lambda_2 \|R_jW^{\top}\bm{d}\|_2,$\\ 
		
		\enspace Return $\bm{d}_1,\ldots, \bm{d}_p$ 
	\end{tabbing}
\end{algorithm}

We also give an algorithm for the extension of our method to classification based on proximal gradient descent. To begin let $L(\bm{d}) = {1}/({2n}) \sum_{i=1}^n \log\left( 1 + \exp\left[ -y_i\left\{\left(RW^{\top} \bm{d}\right)_i\right\}  \right] \right)$, or more generally let it be some differentiable convex loss function. We denote by $\nabla L(\bm{d})$, the derivative of $L$ at the point $\bm{d}\in \mathbb{R}^{\K}$. Algorithm~\ref{alg:univariateLogistic} presents the steps for solving the univariate \name\ problem with general loss. The algorithm for extension of additive models to classification (or other loss functions) can be similarly derived and is omitted in the interest of brevity. 
\begin{algorithm}
	\caption{Proximal gradient descent for extension to classification}
	\label{alg:univariateLogistic}
	\enspace Initialize $\bm{d}^0$ \\
	\enspace For $ l = 1,\, 2,\,\ldots$ until convergence\\
	\qquad Select a step size $t_l$ via line search \\
	\qquad Update $$\bm{d}^{l} \gets \underset{\bm{d} \in \mathbb{R}^{\K}}{\arg\min} \frac{1}{2} \left\| \bm{d}  - \left\{ \bm{d}^{l-1} - t_{l}\nabla L(\bm{d}^{l-1}) \right\}\right\|^2_2 + t_l\lambda \|\bm{d}_{-1}\|_1.$$\\
	
	\enspace Return $\bm{d}^{l}$ 
\end{algorithm}

	% !TeX root = ../MANUSCRIPT.tex

\section{Additional simulation results}
\label{sec:sims}
\begin{figure}
	\centering
	\includegraphics[width=\textwidth]{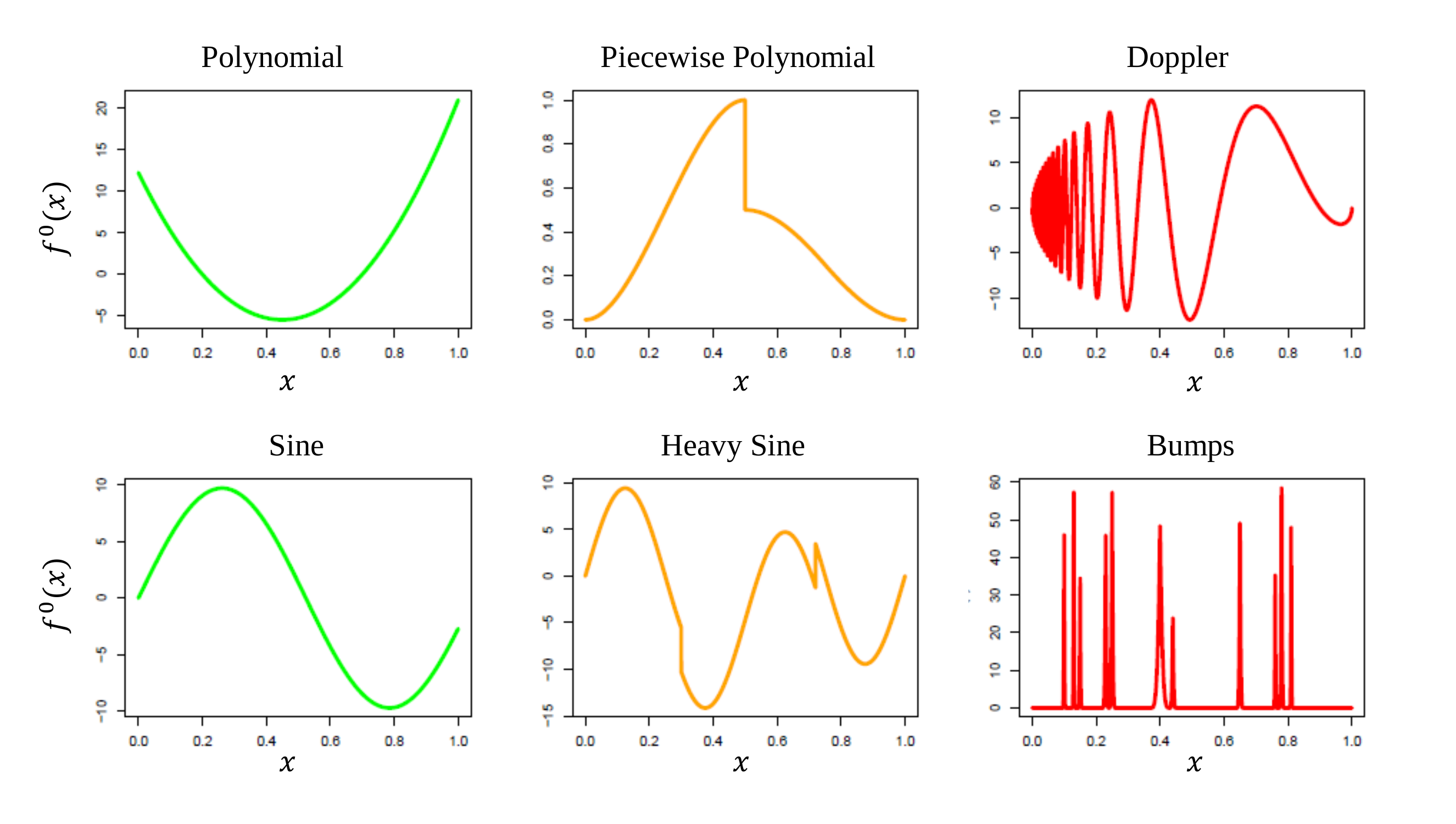}
	\caption{Plots of functions $f^0$ for the simulation study. Functions in green are the most smooth and well-behaved followed by functions with moderate smoothness in orange. Finally, functions in red are highly irregular functions, e.g., functions with unbounded total variation.}
	\label{fig:truef}
\end{figure}

In this section we present some additional simulation results. The simulation study for both univariate and multivariate regression, used six functions: 1. polynomial, 2. sine, 3. piecewise polynomial, 4. heavy sine, 5. bumps and, 6. doppler. The six functions are presented in Figure~\ref{fig:truef}.

\subsection{Univariate simulation study for $x_i\sim \mathcal{N}(0,1)$}
We begin with presenting the table of results for the univariate regression simulation study. In Table~\ref{tab:norm}, we present the results for normally distributed covariates, i.e., $x_i\sim \mathcal{N}(0,1)$, and then scaled to $[0,1]$. We see that other than the polynomial function \name\ generally outperforms competitors in terms of prediction error. 
\begin{table}
	\caption{Table of results for $x_i\sim \mathcal{N}(0, 1)$ averaged over 100 replications of the data. The table presents the ratio $\operatorname{MSE}/\operatorname{MSE}_{FG}$ along with $100\times$ the standard error, where $\operatorname{MSE}_{FG}$ is the $\operatorname{MSE}$ of \name\ with $K = 2^{\lceil \log_2 n\rceil}$. Boldface values represent the method with the smallest MSE within each row of the table.}
	\label{tab:norm}
	\begin{tabular}{llccccc}
		\hline
		& & \name & \name & Interpolation & Isometric & \multicolumn{1}{c}{Adaptive Lifting} \\ 
		& & $K = 2^5$ & $K = 2^6$ &  &  & \\
		\hline
		Polynomial & n = 64  & 1.47 (13.17) & 1.41 (11.32) & \textbf{0.51} (3.04) & 1.45 (10.11) & 1.59 (9.30) \\
		& n = 128  & 0.78 (5.25) & 0.77 (4.95) & \textbf{0.40} (2.96) & 0.87 (4.69) & 0.88 (4.84) \\
		& n = 256  & \textbf{0.39} (3.75) & 0.51 (3.89) & 0.43 (2.38) & 0.64 (2.81) & 0.76 (5.97) \\
		& n = 512  & 0.90 (4.57) & 0.77 (4.03) & \textbf{0.29} (0.98) & 0.43 (1.59) & 0.33 (2.36) \\
		Sine & n = 64  & 0.92 (9.55) & 0.99 (1.49) & 1.48 (11.85) & 2.22 (21.67) & 3.61 (35.07) \\
		& n = 128  & \textbf{0.89} (8.74) & 0.91 (3.77) & 1.71 (10.85) & 1.83 (15.18) & 3.53 (33.07) \\
		& n = 256  & \textbf{0.48} (2.39) & 0.73 (1.53) & 1.48 (8.74) & 1.51 (8.18) & 2.73 (22.25) \\
		& n = 512  & \textbf{0.36} (1.22) & 0.64 (1.63) & 1.03 (5.54) & 0.74 (2.77) & 1.21 (7.62) \\
		\hline
		Piecewise & n = 64  & \textbf{0.78} (1.92) & 0.99 (1.01) & 1.50 (6.50) & 1.64 (7.54) & 2.18 (14.06) \\
		Polynomial & n = 128  & 0.86 (2.29) & \textbf{0.83} (2.04) & 1.89 (7.42) & 1.59 (4.60) & 1.65 (8.86) \\
		& n = 256  & 1.25 (3.80) & \textbf{0.90} (2.22) & 1.64 (5.21) & 1.09 (3.24) & 1.15 (6.94) \\
		& n = 512  & 1.79 (2.71) & 1.27 (2.34) & 1.76 (3.24) & \textbf{0.96} (1.54) & 1.01 (4.29) \\
		Heavy Sine & n = 64  & \textbf{0.73} (1.81) & 1.00 (0.65) & 1.23 (4.40) & 1.26 (4.03) & 1.54 (6.83) \\
		& n = 128  & \textbf{0.54} (1.70) & 0.78 (1.40) & 1.30 (5.02) & 1.14 (2.78) & 1.12 (6.04) \\
		& n = 256  & \textbf{0.47} (0.93) & 0.65 (0.98) & 1.17 (3.08) & 0.89 (1.99) & 0.93 (5.45) \\
		& n = 512  & \textbf{0.38} (0.87) & 0.54 (1.08) & 1.40 (2.91) & 0.77 (1.24) & 0.84 (3.94) \\
		\hline
		Bumps & n = 64  & 1.27 (0.62) & 1.00 (0.06) & 0.85 (1.19) & \textbf{0.36} (0.79) & 0.53 (2.24) \\
		& n = 128  & 3.40 (4.69) & 2.25 (2.81) & 1.35 (2.28) & \textbf{0.69} (1.50) & 0.76 (1.64) \\
		& n = 256  & 6.49 (10.88) & 3.71 (5.58) & 1.31 (2.03) & 1.18 (1.41) & \textbf{1.10} (2.52) \\
		& n = 512  & 8.83 (10.06) & 5.43 (6.03) & 1.29 (1.82) & 1.28 (1.37) & \textbf{1.11} (1.90) \\
		Doppler & n = 64  & \textbf{0.75} (1.84) & 1.00 (0.67) & 1.36 (4.74) & 1.53 (4.32) & 1.56 (6.01) \\
		& n = 128  & 0.99 (1.87) & \textbf{0.81} (1.44) & 1.43 (4.75) & 1.49 (3.81) & 1.40 (4.35) \\
		& n = 256  & 0.58 (1.11) & \textbf{0.52} (1.06) & 1.26 (3.25) & 1.15 (1.86) & 0.98 (3.77) \\
		& n = 512  & 0.98 (1.52) & \textbf{0.58} (1.05) & 1.24 (2.38) & 0.98 (1.48) & 0.85 (2.21) \\
		\hline 
	\end{tabular}
\end{table}

\subsection{Univariate simulation study for sample sizes not a power of two}

In this section, we present results for the simulation study of Section 4 for sample sizes $n = 75, 100, 300, 500$. The results are presented for $x_i\sim \mathcal{U}(0,1)$ and $x_i\sim \mathcal{N}(0,1)$ in Table~\ref{tab:unif2} and \ref{tab:norm2}, respectively. 

\begin{table}
		\caption{Table of results for $x_i\sim \mathcal{U}[0, 1]$ averaged over 100 replications of the data for sample sizes that are not powers of 2. The table presents the ratio $\operatorname{MSE}/\operatorname{MSE}_{FG}$ along with $100\times$ the standard error, where $\operatorname{MSE}_{FG}$ is the $\operatorname{MSE}$ of \name\ with $K = 2^{\lceil \log_2 n\rceil}$. Boldface values represent the method with the smallest MSE within each row of the table.}
	\label{tab:unif2}
\begin{tabular}{llccccc}
	\hline
	& & \name & \name & Interpolation & Isometric & \multicolumn{1}{c}{Adaptive Lifting} \\ 
	& & $K = 2^5$ & $K = 2^6$ &  &  & \\
	\hline
	Polynomial & n = 75  & 1.20 (3.51) & \textbf{1.00} (0.00) & 1.32 (3.71) & 4.98 (22.10) & 4.35 (24.12) \\
	& n = 100  & 1.18 (3.96) & \textbf{1.00} (0.00) & 1.39 (4.76) & 4.24 (17.06) & 3.98 (21.99) \\
	& n = 300  & 0.84 (3.02) & \textbf{0.81} (2.55) & 1.87 (5.91) & 5.33 (18.57) & 3.86 (19.36) \\
	& n = 500  & 0.96 (2.95) & \textbf{0.89} (2.69) & 2.13 (5.91) & 3.36 (14.51) & 4.21 (21.17) \\
	Sine & n = 75  & 1.09 (3.76) & \textbf{1.00} (0.00) & 1.55 (5.52) & 2.57 (11.78) & 3.78 (21.66) \\
	& n = 100  & 1.04 (2.73) & \textbf{1.00} (0.00) & 1.67 (6.81) & 1.75 (5.65) & 3.43 (19.26) \\
	& n = 300  & \textbf{0.67} (1.77) & 0.73 (1.95) & 2.33 (7.06) & 2.18 (6.65) & 4.26 (25.07) \\
	& n = 500  & \textbf{0.73} (2.08) & 0.76 (2.42) & 2.72 (8.77) & 1.28 (3.35) & 4.05 (21.45) \\
	\hline
	Piecewise & n = 75  & \textbf{0.87} (1.55) & 1.00 (0.00) & 1.32 (2.60) & 1.40 (3.16) & 1.73 (7.43) \\
	Polynomial & n = 100  & \textbf{0.84} (1.42) & 1.00 (0.00) & 1.33 (2.93) & 1.39 (2.59) & 1.40 (5.63) \\
	& n = 300  & 0.98 (1.47) & \textbf{0.92} (1.25) & 1.63 (2.51) & 1.27 (1.68) & 1.40 (5.28) \\
	& n = 500  & 1.19 (1.57) & \textbf{1.03} (1.13) & 1.95 (3.23) & 1.26 (1.48) & 1.36 (4.01) \\
	Heavy Sine & n = 75  & \textbf{0.89} (1.87) & 1.00 (0.00) & 1.31 (2.92) & 1.44 (3.01) & 1.79 (6.43) \\
	& n = 100  & \textbf{0.87} (1.49) & 1.00 (0.00) & 1.38 (2.92) & 1.72 (3.50) & 1.59 (5.33) \\
	& n = 300  & \textbf{0.73} (1.39) & 0.81 (1.01) & 1.87 (3.03) & 1.87 (2.93) & 1.80 (5.81) \\
	& n = 500  & \textbf{0.76} (1.23) & 0.80 (1.04) & 1.99 (3.17) & 1.61 (1.97) & 1.77 (5.37) \\
	\hline
	Bumps & n = 75  & 1.83 (1.10) & 1.00 (0.00) & 0.76 (1.00) & \textbf{0.46} (0.61) & 0.88 (3.90) \\
	& n = 100  & 1.56 (0.59) & 1.00 (0.00) & 0.72 (0.80) & \textbf{0.38} (0.48) & 0.61 (1.75) \\
	& n = 300  & 4.47 (3.00) & 3.20 (1.99) & 0.87 (0.54) & \textbf{0.81} (0.60) & 0.83 (1.17) \\
	& n = 500  & 4.57 (2.03) & 3.51 (1.48) & 0.80 (0.52) & \textbf{0.74} (0.53) & 0.74 (0.73) \\
	Doppler & n = 75  & \textbf{0.96} (1.39) & 1.00 (0.00) & 1.19 (1.95) & 1.47 (2.54) & 1.40 (3.54) \\
	& n = 100  & 1.18 (1.30) & \textbf{1.00} (0.00) & 1.25 (2.13) & 1.50 (2.33) & 1.37 (2.97) \\
	& n = 300  & 2.27 (3.46) & \textbf{1.10} (1.15) & 1.36 (1.53) & 1.36 (1.51) & 1.37 (2.30) \\
	& n = 500  & 3.44 (4.69) & 1.70 (1.98) & 1.60 (1.69) & \textbf{1.42} (1.60) & 1.59 (2.29) \\
	\hline 
\end{tabular}
	
\end{table}

\begin{table}
		\caption{Table of results for $x_i\sim \mathcal{N}(0, 1)$ averaged over 100 replications of the data for sample sizes that are not powers of 2. The table presents the ratio $\operatorname{MSE}/\operatorname{MSE}_{FG}$ along with $100\times$ the standard error, where $\operatorname{MSE}_{FG}$ is the $\operatorname{MSE}$ of \name\ with $K = 2^{\lceil \log_2 n\rceil}$. Boldface values represent the method with the smallest MSE within each row of the table.}
\label{tab:norm2}
\begin{tabular}{llccccc}
	\hline
	& & \name & \name & Interpolation & Isometric & \multicolumn{1}{c}{Adaptive Lifting} \\ 
& & $K = 2^5$ & $K = 2^6$ &  &  & \\
	\hline
	Polynomial & n = 75  & 1.11 (3.83) & 1.00 (0.00) & \textbf{0.99} (4.31) & 3.93 (18.03) & 3.10 (21.66) \\
	& n = 100  & 1.26 (4.25) & \textbf{1.00} (0.00) & 1.16 (4.45) & 4.24 (21.09) & 2.94 (17.20) \\
	& n = 300  & 0.68 (2.77) & \textbf{0.49} (1.11) & 1.01 (3.27) & 3.19 (7.76) & 1.74 (8.14) \\
	& n = 500  & 0.37 (0.93) & \textbf{0.37} (0.77) & 0.80 (2.01) & 1.57 (2.39) & 0.92 (4.36) \\
	Sine & n = 75  & \textbf{0.86} (2.86) & 1.00 (0.00) & 1.16 (5.27) & 1.81 (6.86) & 2.64 (12.89) \\
	& n = 100  & \textbf{0.87} (2.66) & 1.00 (0.00) & 1.24 (5.00) & 1.82 (6.40) & 2.49 (13.27) \\
	& n = 300  & \textbf{0.67} (1.75) & 0.86 (1.53) & 1.12 (4.43) & 1.63 (5.41) & 1.77 (8.27) \\
	& n = 500  & 0.74 (2.10) & \textbf{0.73} (1.50) & 1.41 (4.43) & 1.23 (3.16) & 1.69 (7.46) \\
	\hline
	Piecewise & n = 75  & \textbf{0.95} (1.90) & 1.00 (0.00) & 1.54 (4.27) & 1.65 (4.39) & 1.74 (7.56) \\
	Polynomial & n = 100  & \textbf{0.96} (1.95) & 1.00 (0.00) & 1.70 (4.76) & 1.54 (4.35) & 1.55 (6.54) \\
	& n = 300  & 1.32 (2.34) & \textbf{0.91} (1.07) & 1.76 (3.67) & 1.27 (2.43) & 1.25 (4.37) \\
	& n = 500  & 1.76 (2.45) & 1.20 (1.41) & 1.79 (3.23) & \textbf{0.95} (1.54) & 1.03 (3.29) \\
	Heavy Sine & n = 75  & \textbf{0.81} (1.47) & 1.00 (0.00) & 1.25 (3.14) & 1.48 (2.68) & 1.56 (5.69) \\
	& n = 100  & \textbf{0.85} (2.11) & 1.00 (0.00) & 1.48 (4.03) & 1.68 (3.34) & 1.47 (5.58) \\
	& n = 300  & \textbf{0.55} (1.34) & 0.66 (1.14) & 1.33 (2.59) & 1.47 (1.93) & 1.01 (3.28) \\
	& n = 500  & 0.73 (1.85) & \textbf{0.72} (1.12) & 1.71 (2.80) & 1.12 (1.47) & 1.03 (3.56) \\
	\hline
	Bumps & n = 75  & 1.28 (0.43) & 1.00 (0.00) & 0.69 (0.59) & \textbf{0.34} (0.54) & 0.48 (1.66) \\
	& n = 100  & 1.42 (0.52) & 1.00 (0.00) & 0.65 (0.56) & \textbf{0.35} (0.49) & 0.41 (1.34) \\
	& n = 300  & 4.58 (3.42) & 3.04 (2.07) & 0.83 (0.84) & 0.83 (0.76) & \textbf{0.81} (0.98) \\
	& n = 500  & 7.20 (5.39) & 4.46 (3.24) & 1.08 (1.03) & 1.10 (1.01) & \textbf{0.91} (1.11) \\
	Doppler & n = 75  & 1.09 (1.70) & \textbf{1.00} (0.00) & 1.37 (3.22) & 1.63 (3.08) & 1.58 (3.96) \\
	& n = 100  & 1.23 (1.69) & \textbf{1.00} (0.00) & 1.46 (3.17) & 1.77 (3.39) & 1.61 (4.31) \\
	& n = 300  & 0.68 (1.29) & \textbf{0.66} (1.11) & 1.67 (2.59) & 1.51 (2.34) & 1.21 (2.99) \\
	& n = 500  & 1.36 (1.69) & \textbf{0.82} (0.77) & 1.79 (2.63) & 1.35 (1.78) & 1.22 (2.61) \\
	\hline 
\end{tabular}	
\end{table}

\subsection{Effect of truncation level $K$}
\begin{figure}
	\includegraphics[scale = 0.45,page=1]{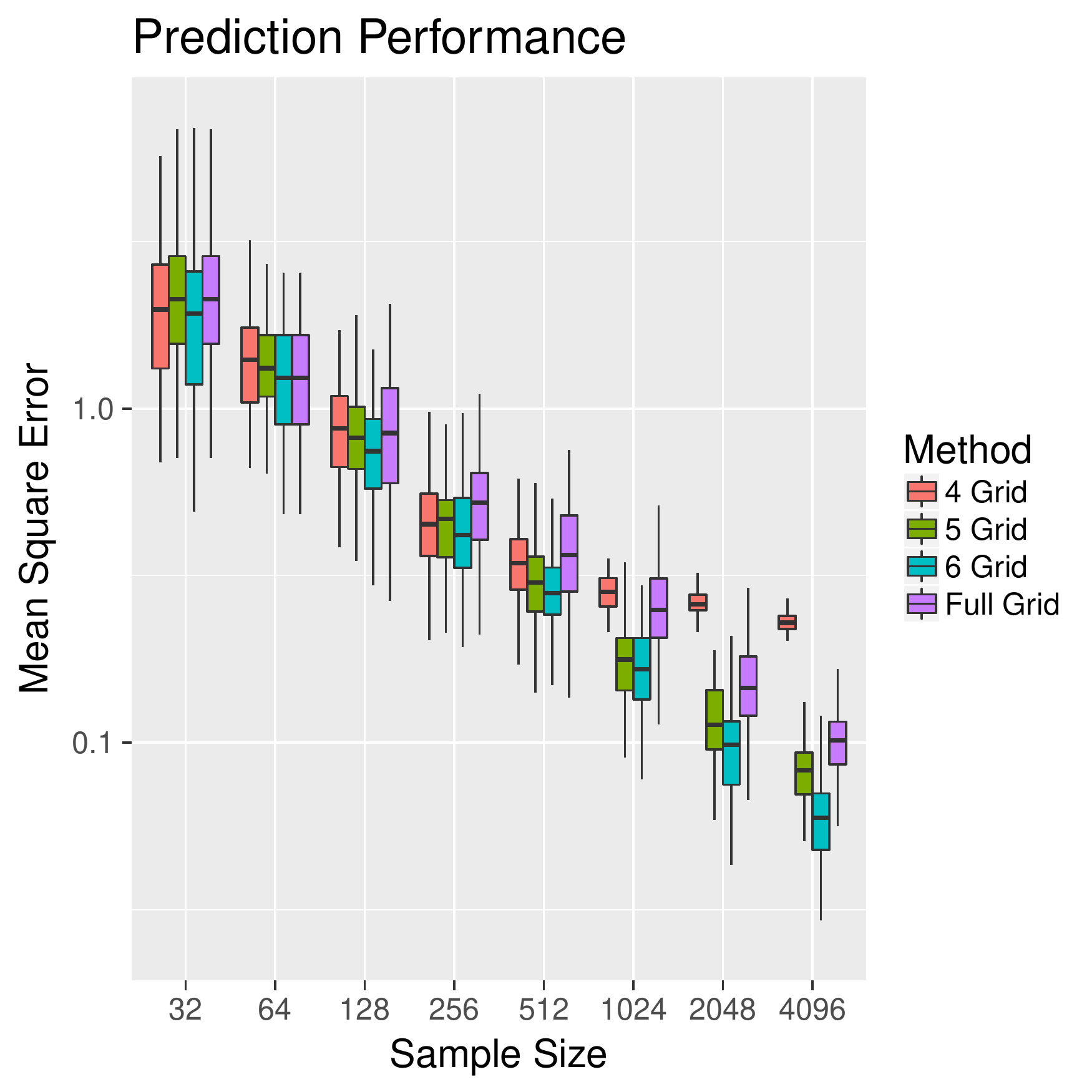}
	\includegraphics[scale = 0.45,page=2]{plots/Polynomial.pdf}
		\caption{Effect of truncation level $K$. Results of for the Polynomial function.}
			\label{fig:poly}
\end{figure}

\begin{figure}
	\includegraphics[scale = 0.45,page=1]{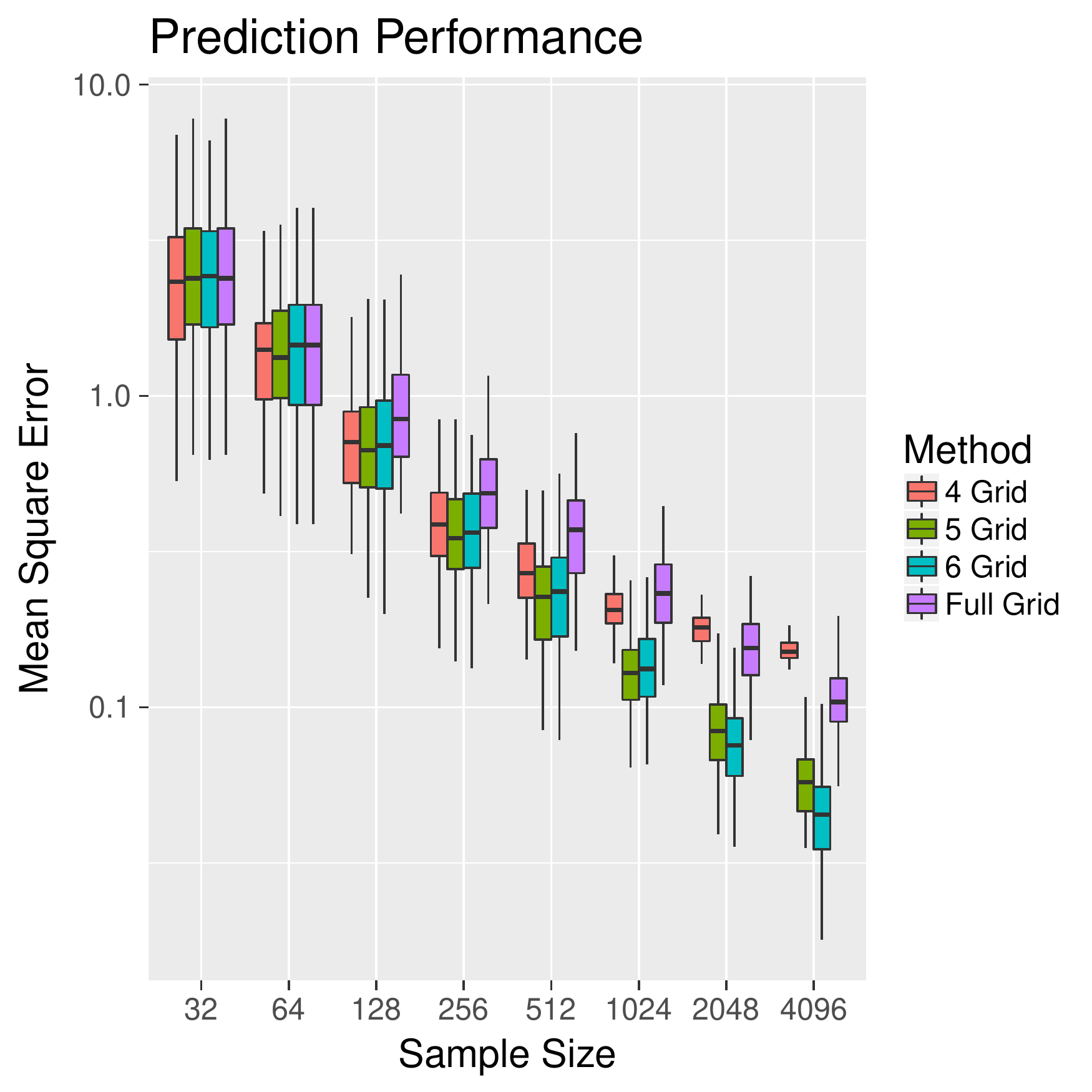}
	\includegraphics[scale = 0.45,page=2]{plots/Sine.pdf}
	\caption{Effect of truncation level $K$. Results of for the Sine function.}
		\label{fig:sin}
		
\end{figure}

\begin{figure}
	\includegraphics[scale = 0.45,page=1]{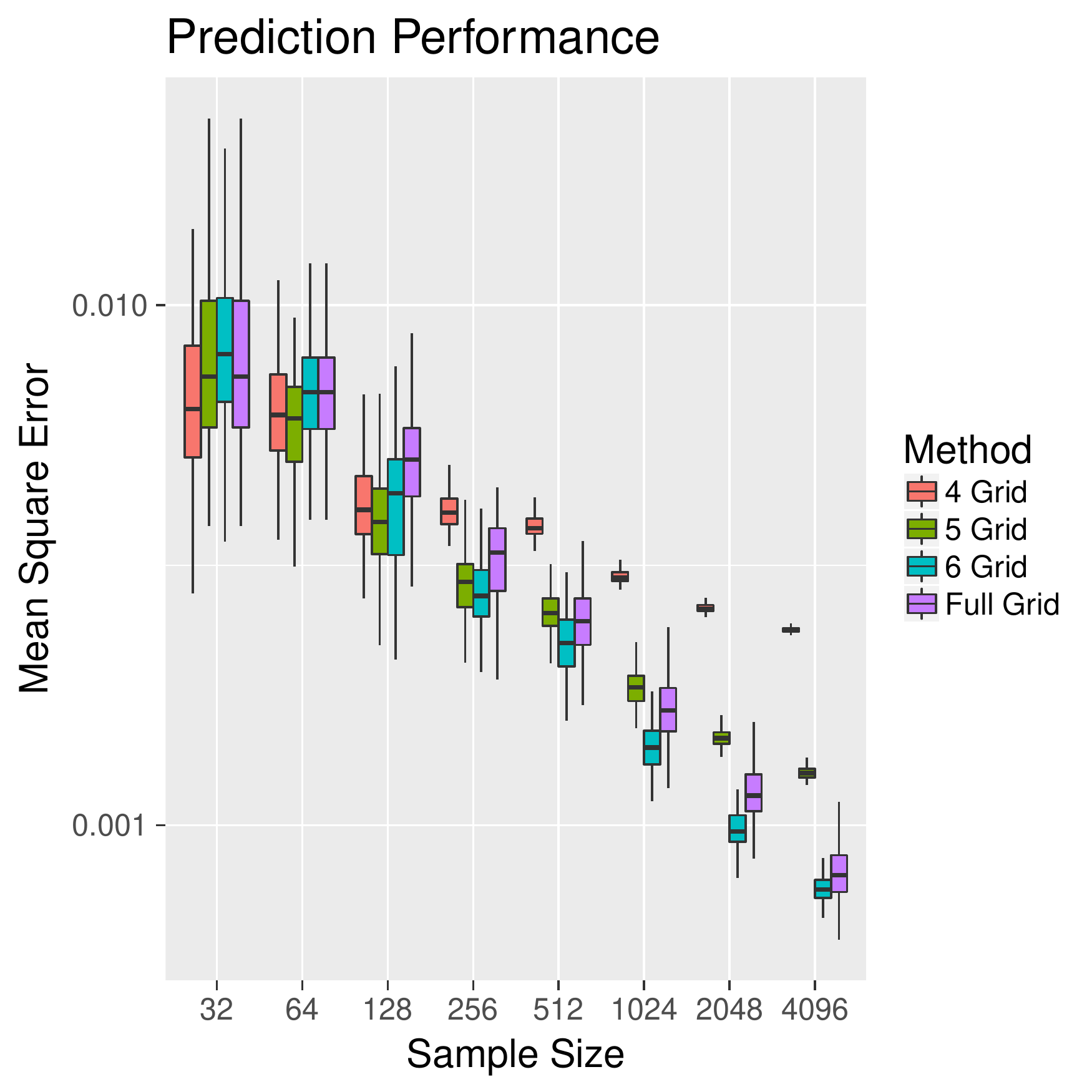}
	\includegraphics[scale = 0.45,page=2]{plots/PiecewisePolynomial.pdf}
	\caption{Effect of truncation level $K$. Results of for the Piecewise Polynomial function.}
	\label{fig:ppoly}
\end{figure}

\begin{figure}
	\includegraphics[scale = 0.45,page=1]{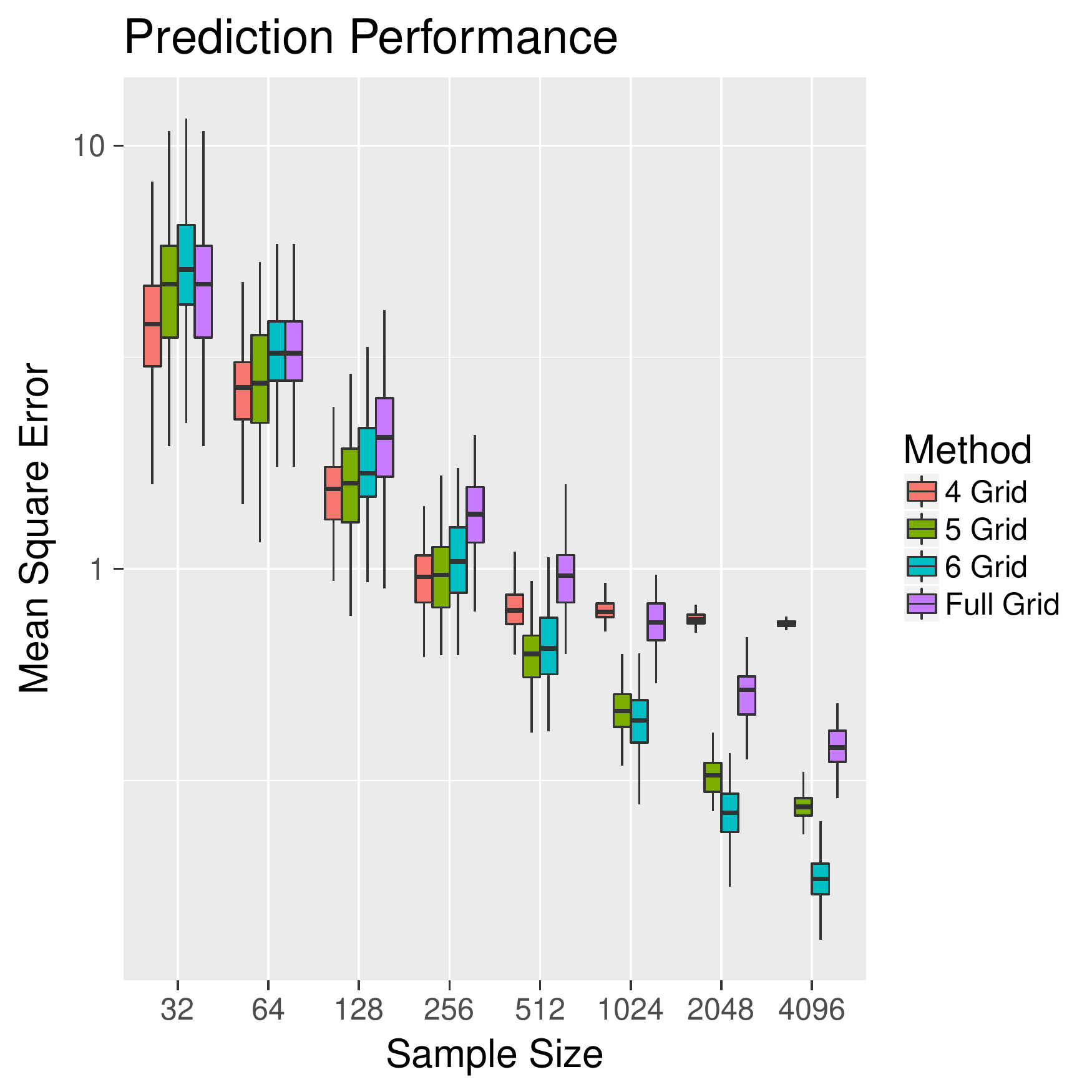}
	\includegraphics[scale = 0.45,page=2]{plots/Heaviside.pdf}
	\caption{Effect of truncation level $K$. Results of for the Heavy sine function.}
	\label{fig:heavi}
\end{figure}

\begin{figure}
	\includegraphics[scale = 0.45,page=1]{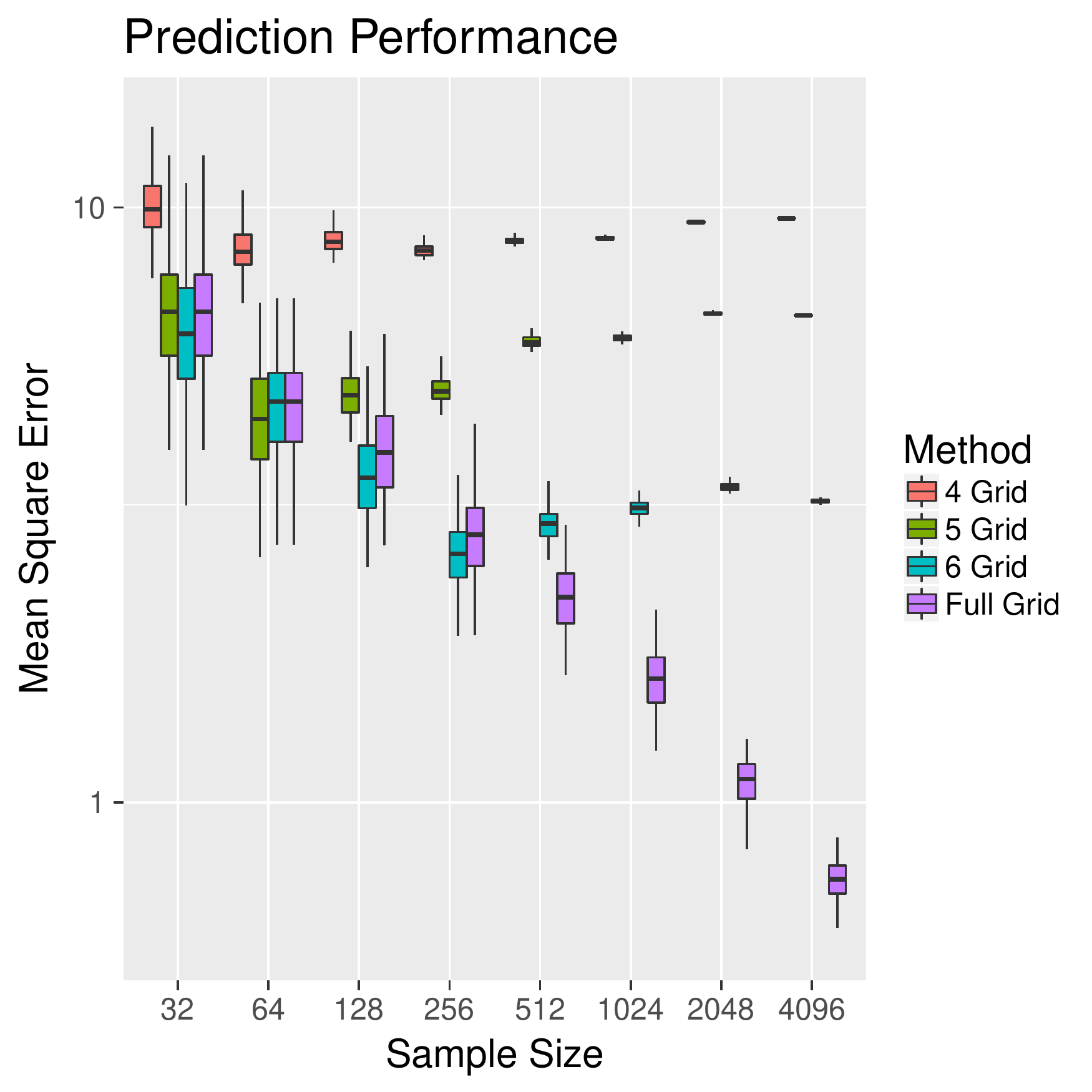}
	\includegraphics[scale = 0.45,page=2]{plots/Doppler.pdf}
	\caption{Effect of truncation level $K$. Results of for the Doppler function.}
	\label{fig:doppler}
\end{figure}

\begin{figure}
	\includegraphics[scale = 0.45,page=1]{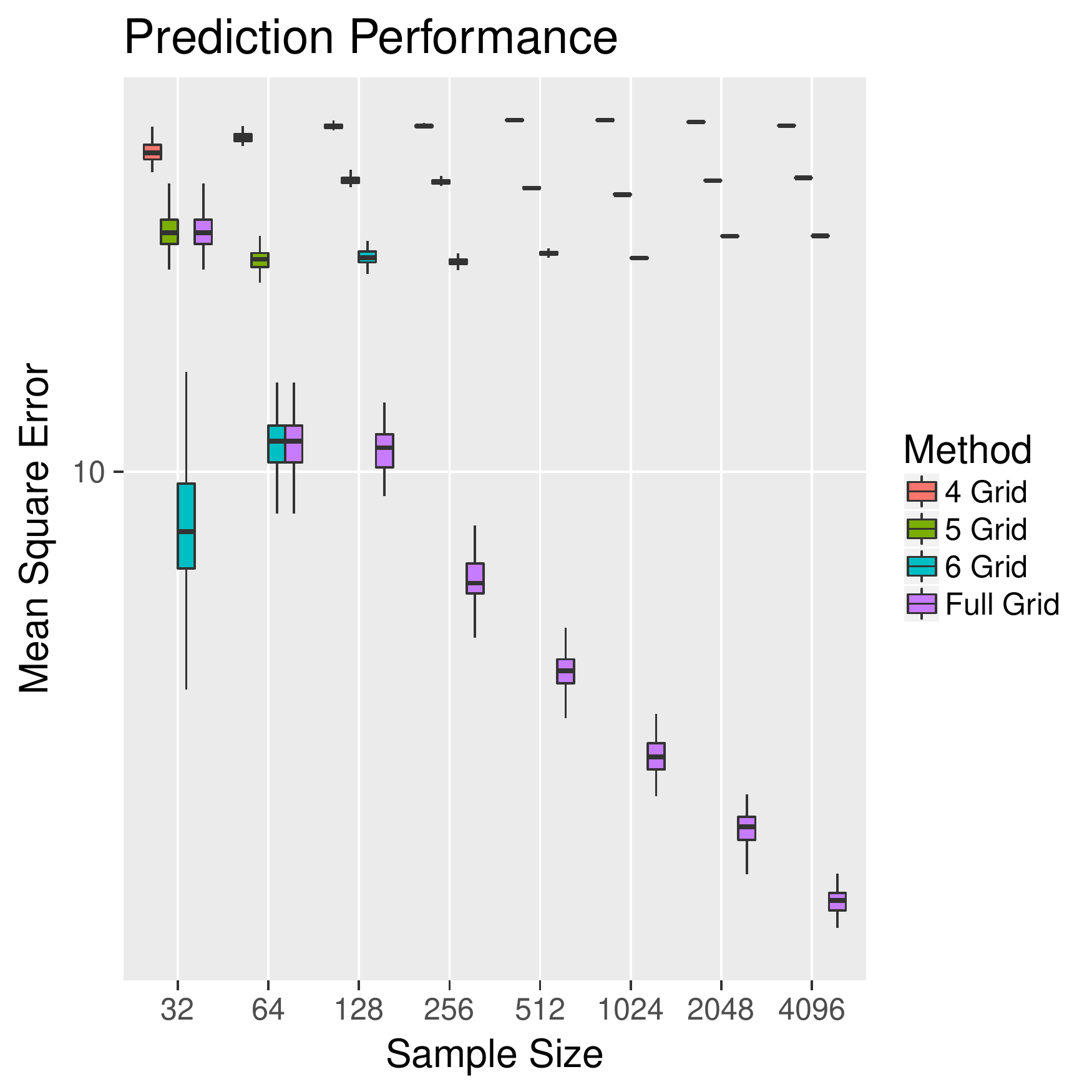}
	\includegraphics[scale = 0.45,page=2]{plots/Bumps.pdf}
	\caption{Effect of truncation level $K$. Results of for the Bumps function.}
	\label{fig:bumps}
\end{figure}

In this subsection, we present simulation results which study the effects of using different truncation levels $K$. In Figures~\ref{fig:poly} to \ref{fig:bumps} we plot the results for each of the 6 functions considered in the simulation of the manuscript. 

In the left panel of each figure we plot the MSE as a function of sample size, $n$. This is done for the full grid method where we take $K = 2^{\log_2 n}$, and for \name\ with $K = 2^4, 2^5$ and $2^6$ which we refer to as 4 Grid, 5 Grid and 6 Grid, respectively. In the right panel of each figure we present the computation time as a function of sample size $n$ for \name\ with $K = 2^4, 2^5, 2^6$ and $2^{\log_2 n}$. 

We see in Figures \ref{fig:doppler} and \ref{fig:bumps}, that using a small order $K$ leads to substantially high MSE. This is most likely due to the nature of the underlying functions. The Doppler function is an example of function which does not have a bounded variation, estimating such functions by interpolation is extremely difficult and in general we need a full grid, i.e. $K=n$. On the other hand for all other functions, i.e. polynomial, sine etc, we see a clear advantage of using $K = 2^7$ basis functions. We also see in some figures that while using $K = 2^6$ leads to substantially smaller MSE using too small a value of $K$ can be lead to poor prediction performance. We see this even in the simple cases of estimating a polynomial or sine function. 

We notice on the right panels the clear computational advantage of using fewer than $n$ basis functions. We observe the computation time for fixed $K$ generally does not vary too much with increasing sample size. This is because the main computational step is the DWT and IDWT via Mallets algorithm. The other matrix multiplications are sparse and can be computed efficiently.

\subsection{Simulation study for adaptive \name}

\begin{figure}
	\includegraphics[scale = 0.45,page=1]{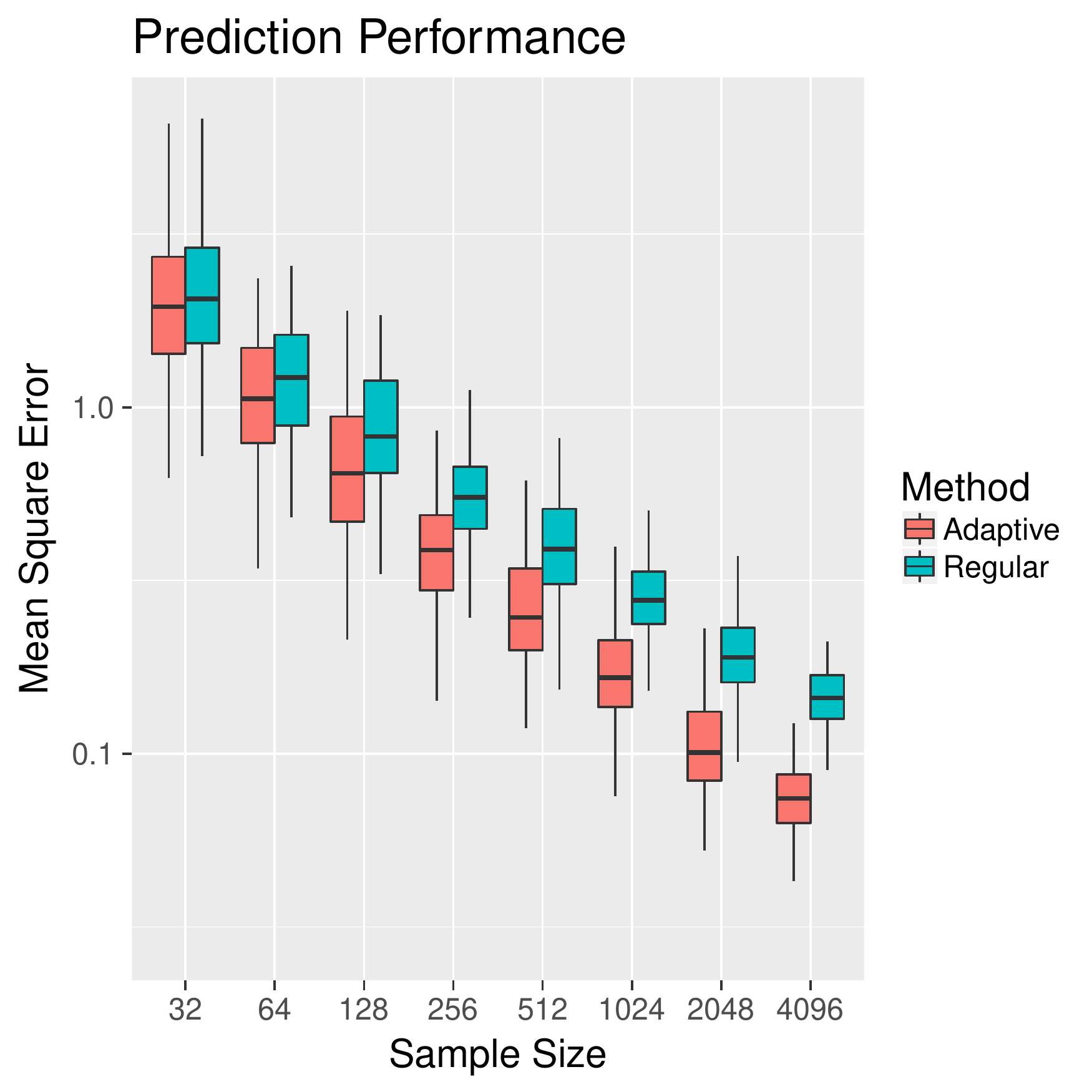}
	\includegraphics[scale = 0.45,page=2]{plotsAdap/Polynomial.pdf}\\
	\caption{Simulation study for adaptive \name. Results of for the Polynomial function.}
		\label{fig:poly2}
		
\end{figure}

\begin{figure}
	\includegraphics[scale = 0.45,page=1]{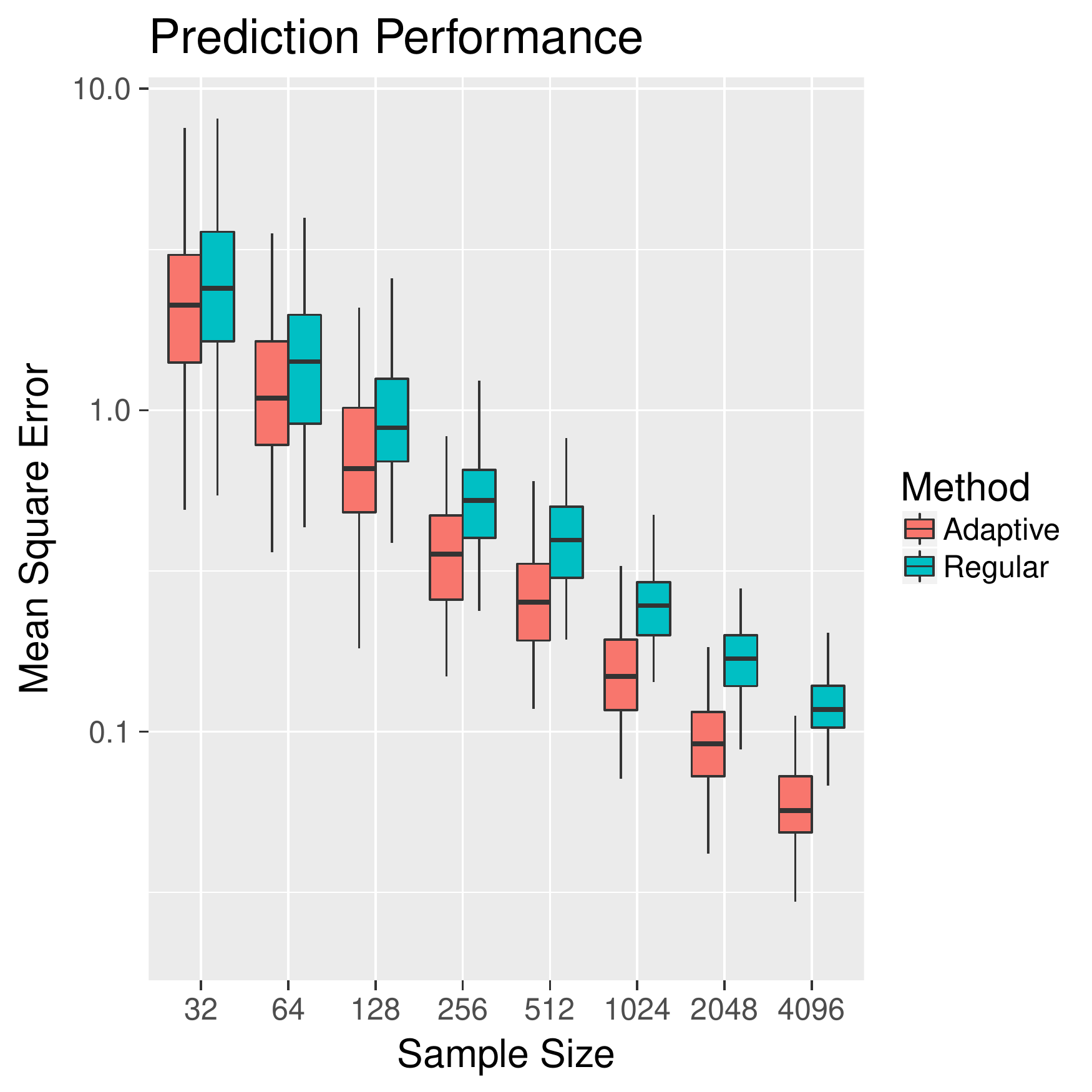}
	\includegraphics[scale = 0.45,page=2]{plotsAdap/Sine.pdf}
	\caption{Simulation study for adaptive \name. Results of for the Sine function.}
		\label{fig:sin2}
		
\end{figure}

\begin{figure}
	\includegraphics[scale = 0.45,page=1]{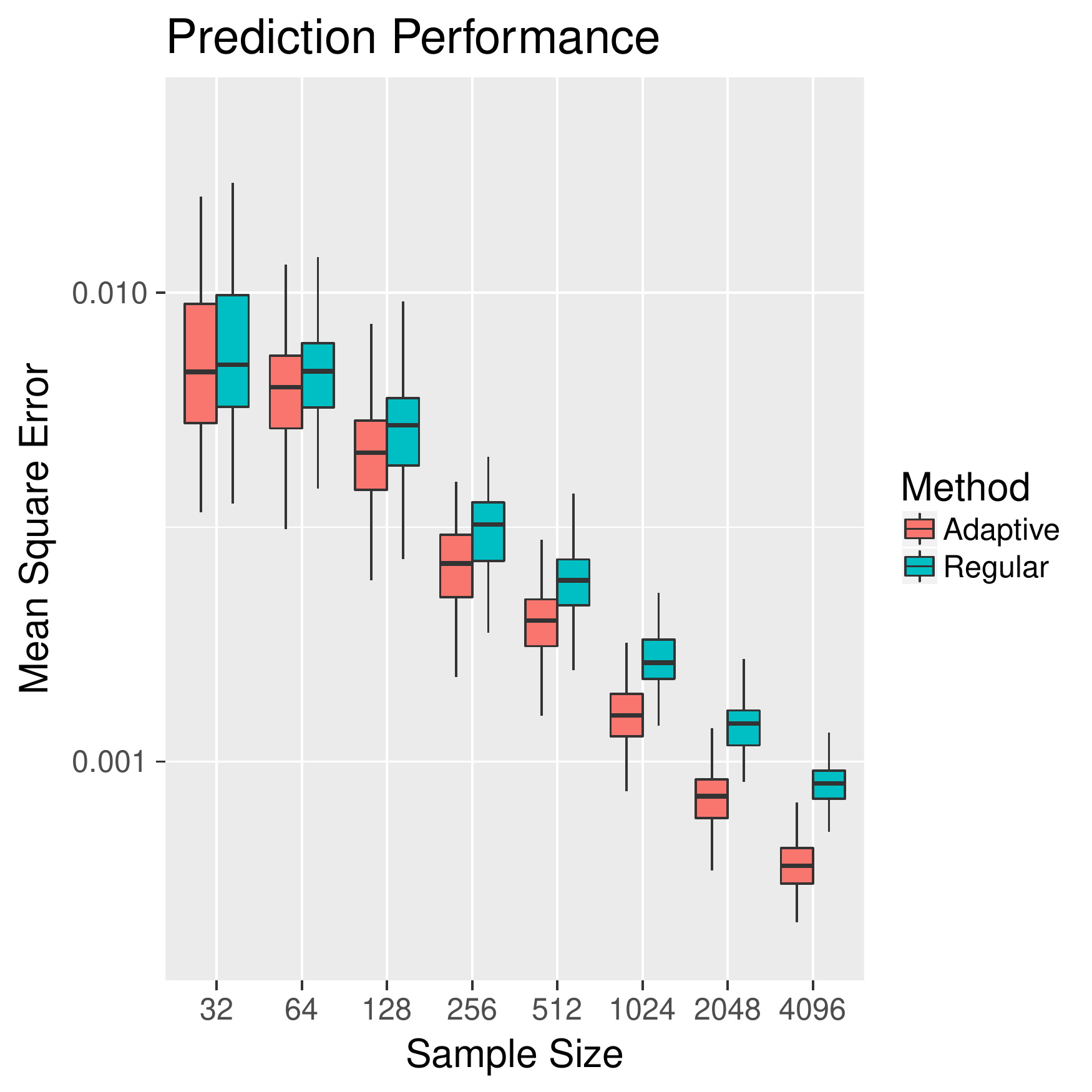}
	\includegraphics[scale = 0.45,page=2]{plotsAdap/PiecewisePolynomial.pdf}
	\caption{Simulation study for adaptive \name. Results of for the Piecewise Polynomial function.}
		\label{fig:ppoly2}
		
\end{figure}

\begin{figure}
	\includegraphics[scale = 0.45,page=1]{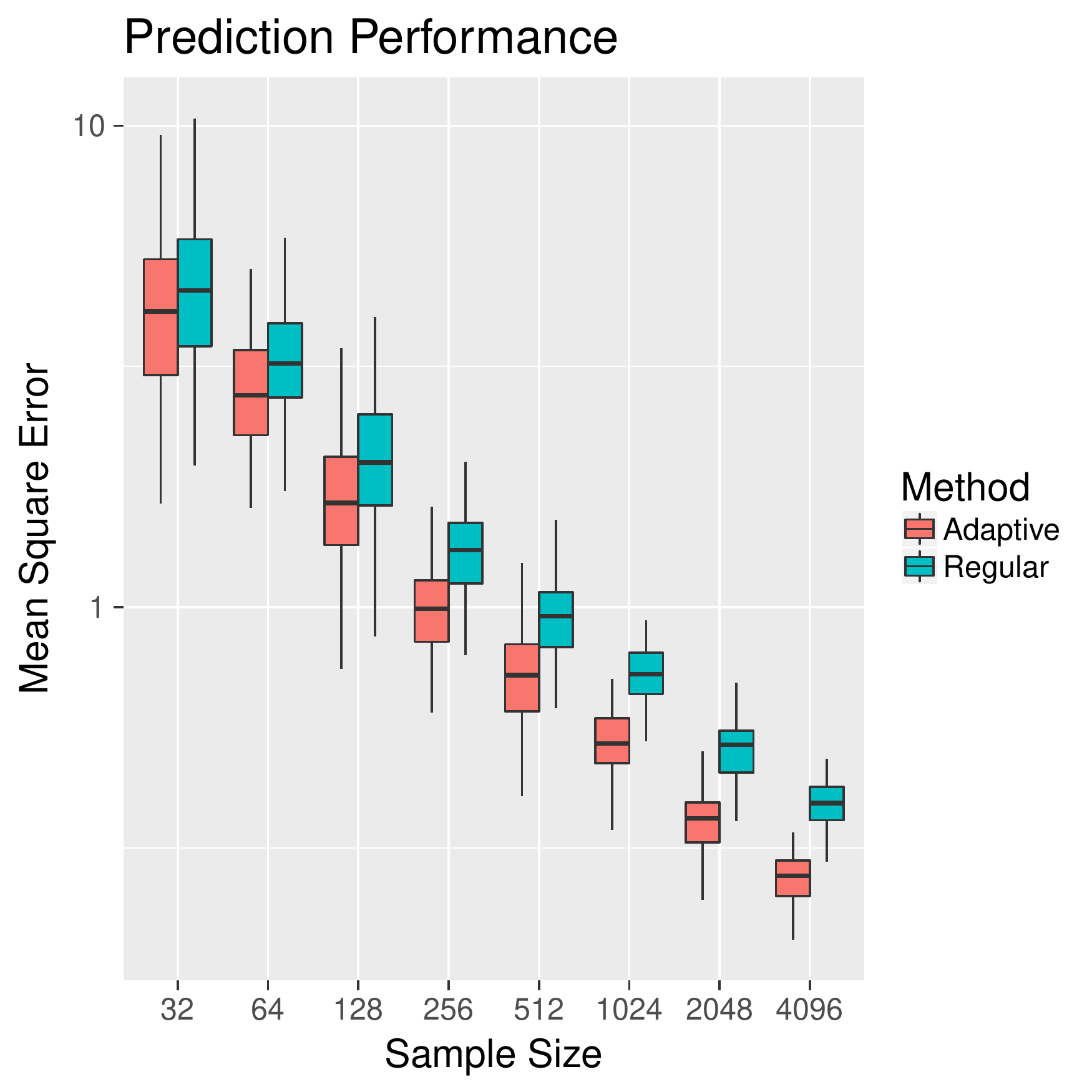}
	\includegraphics[scale = 0.45,page=2]{plotsAdap/Heaviside.pdf}
	\caption{Simulation study for adaptive \name. Results of for the Heavy sine function.}
		\label{fig:heavi2}
\end{figure}

\begin{figure}
	\includegraphics[scale = 0.45,page=1]{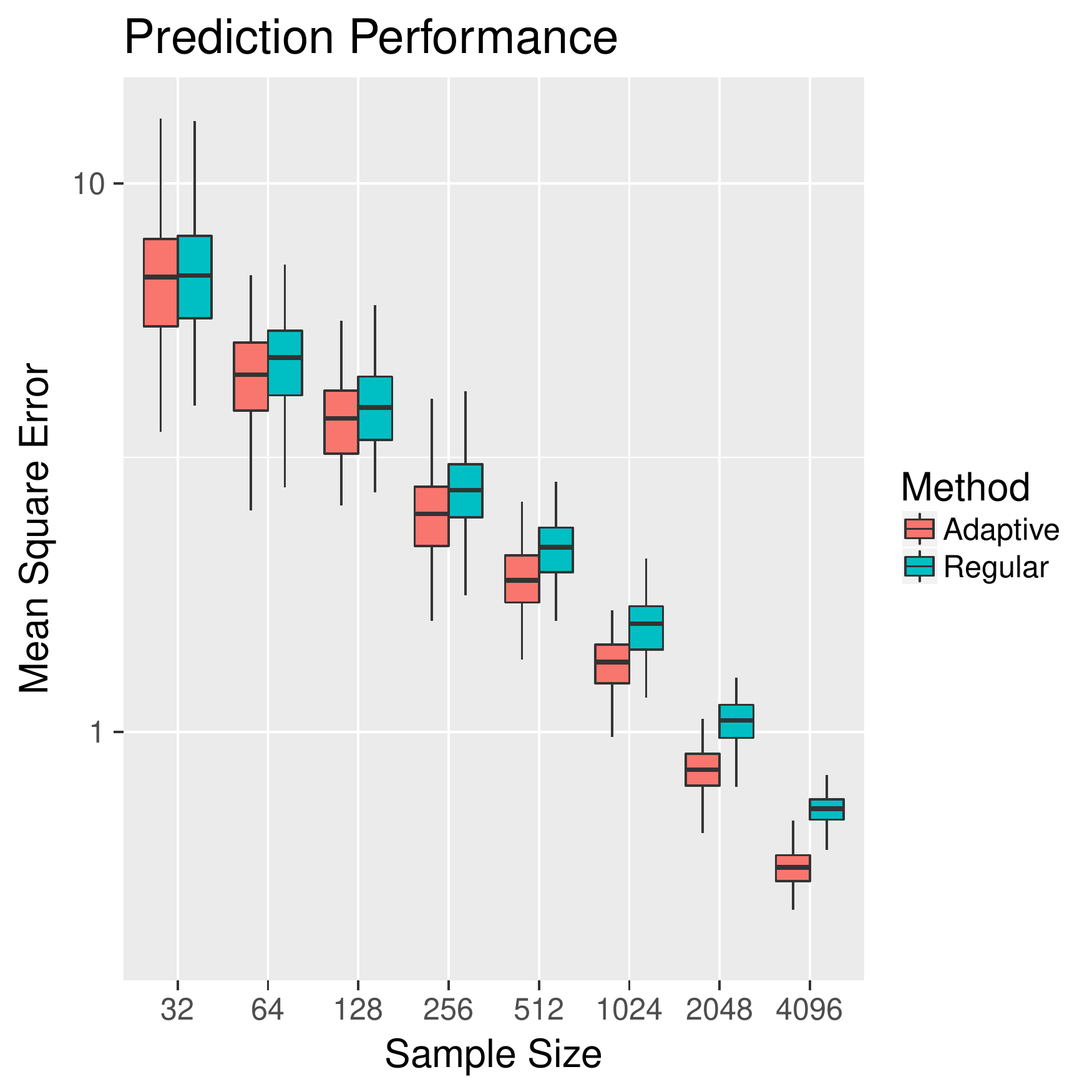}
	\includegraphics[scale = 0.45,page=2]{plotsAdap/Doppler.pdf}
	\caption{Simulation study for adaptive \name. Results of for the Doppler function.}
		\label{fig:doppler2}
\end{figure}

\begin{figure}
	\includegraphics[scale = 0.45,page=1]{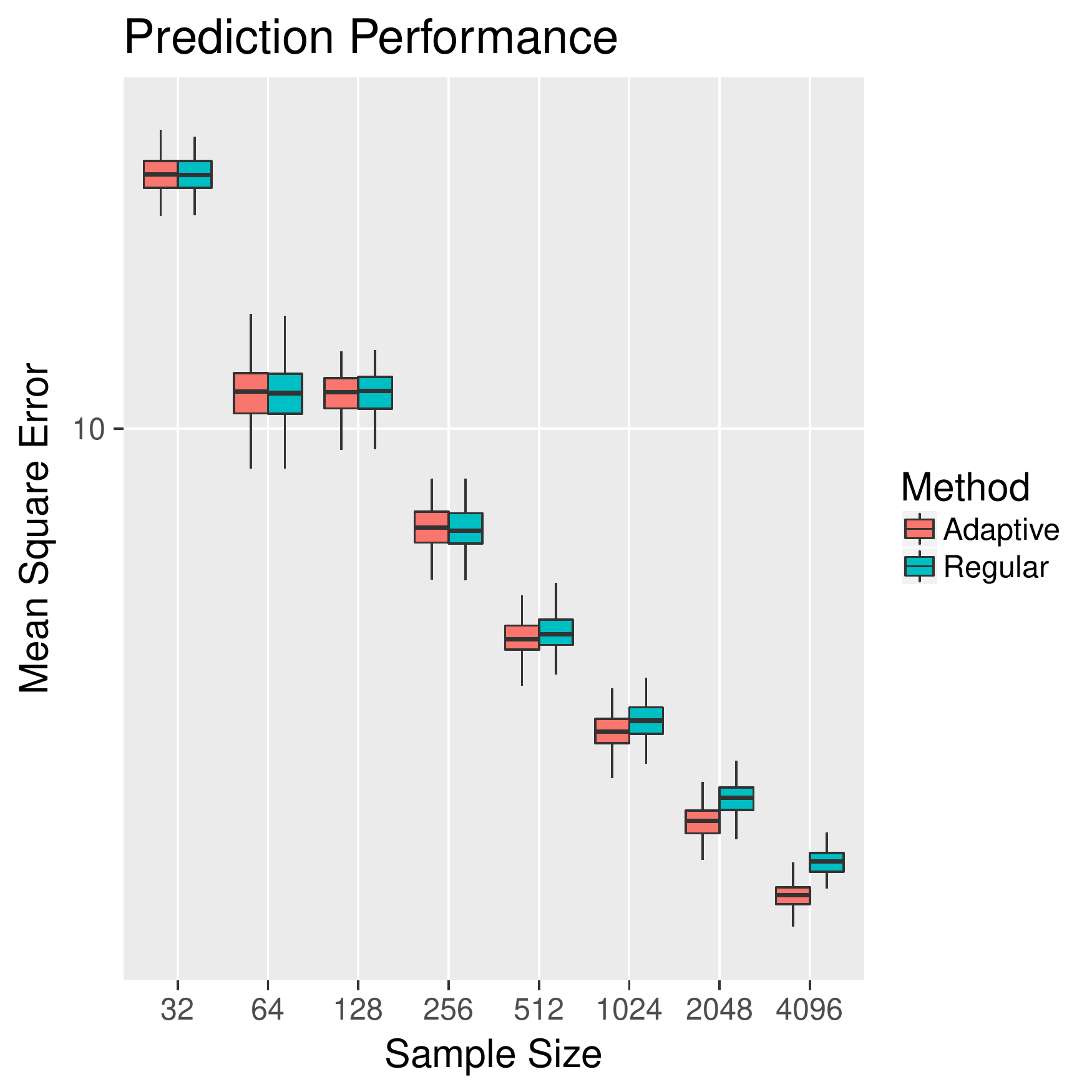}
	\includegraphics[scale = 0.45,page=2]{plotsAdap/Bumps.pdf}
	\caption{Simulation study for adaptive \name. Results of for the Bumps function.}
		\label{fig:bumps2}
\end{figure}

Finally, in this subsection, we present some simulation results regarding the adaptive \name\ estimator introduced in Section 2.4 of the Manuscript. In the left panel of Figure~\ref{fig:poly2} to \ref{fig:bumps2} we present the MSE as a function of sample size for regular \name\ with $K =n$ and adaptive \name. We present the minimum MSE over a sequence of 50 $\lambda$ values. We see that our adaptive estimator uniformly outperforms the regular estimator in terms of prediction error. The results indicates that if we have a good procedure for selecting the tuning parameter, i.e., if we pick close to the theoretically ideal tuning parameter then adaptive \name\ will have a lower MSE.

	\newpage
	
	% !TeX root = ../MANUSCRIPT.tex

\section{Proofs for univariate results}
\label{sec:proofs}

%\subsection{Proofs for univariate regression}
Here we present the proof for Theorem 1. We consider the estimator  
\begin{equation}
\label{eqn:waveMesh}
	\wh{\bm{d}} \gets \underset{\bm{d} \in \mathbb{R}^K }{\arg\min} \frac{1}{2n}\|\bm{y} - RW^{\top}\bm{d}\|_2^2 + \lambda\|\bm{d}_M\|_1,
\end{equation}
where $\bm{d}_M$ denotes the sub-vector corresponding to the mother wavelet coefficients. We use this notation to generalize the case of $j_0=0$ where $j_0$ denotes the minimum resolution level. One nice feature about \eqref{eqn:waveMesh} is that it is exactly the lasso problem~\citep{tibshirani1996regression} with design matrix $RW^{\top}$.

\begin{proof}[Proof of Theorem 1]
	We can divide the proof into three parts, (1) the deterministic part, (2) the stochastic part and (3) the approximation error part. The first 2 parts are standard in the lasso literature, for this reason we will use the results from the book by \cite{geer2016estimation}. 
	
	\textbf{Deterministic Part}

	As per Theorem 2.1 of \cite{geer2016estimation} let $\lambda_{\e}$ satisfy 
	\begin{equation*}
		\lambda_{\e} \ge \|WR^{\top}\bm{\e}\|_{\infty}/n,
	\end{equation*}
	where $\bm{\e}$ is the noise vector. Define for $\lambda>\lambda_{\e}$
	\begin{equation*}
		\overline{\lambda} = \lambda + \lambda_{\e}, \quad \underline{\lambda} = \lambda - \lambda_{\e}, 
	\end{equation*}
	and stretching factor $L = \overline{\lambda}/\underline{\lambda}$. Further more, for an index set $S\subset \{1,\ldots, K\}$ and stretching factor $L$ define the \emph{compatibility constant} as
	\begin{equation}
		\vt^2(L,S) = \min \left\{ n^{-1}|S|\|RW^{\top}\bm{d}\|_2^2 :\|\bm{d}_{S}\|_1 = 1, \|\bm{d}_{-S}\|_1\le L \right\},
	\end{equation}
	where $\bm{d}_S$ is the vector $\bm{d}$ with values equal to 0 for indices in $S$. Similarly $\bm{d}_{-S}$ is the vector $\bm{d}$
 with values equal to 0 for indices in $S^c$. Then we have for any set $S$, and vector $\bm{d}^*$ we have
 \begin{equation}
 n^{-1}\|\wh{\bm{f}} - \bm{f}^0\|_2^2 \le n^{-1}\|\bm{f}^0 - RW^{\top}\bm{d}^*\|_2^2 + \frac{|S|\overline{\lambda}^2}{\vt^2(L, S)}.
 \end{equation}	
 For simplicity we take the $\lambda = 2\lambda_{\e}$ giving us $\overline{\lambda} = 3\lambda_{\e}$, $\underline{\lambda} = \lambda_{\e}$ and $L = 3$.
\end{proof}
We consider a quick calculation of the compatibility constant $\vt^(L,S)$. Let $\Lambda_{\min}(R)$ be the minimum eigenvalue of $R$, this will normally be greater than 0 if $K<n$. We then note that:
\begin{align*}
	n^{-1}|S|\|RW^{\top}\bm{d}\|_2^2 &\ge \Lambda_{\min}(R)n^{-1}|S| \|\bm{d}\|_2^2\\
	&= \Lambda_{\min}(R) n^{-1} |S| \left\{ \|\bm{d}_S\|_2^2 + \|\bm{d}_{-S}\|_2^2 \right\}\\
	&\ge \Lambda_{\min}(R) n^{-1} |S| \left\{ \frac{\|\bm{d}_S\|^2_1}{|S|} + \frac{\|\bm{d}_{-S}\|_1^2}{K - |S|} \right\},
\end{align*}
and minimizing the right hand side under the constraints $\|\bm{d}_{S}\|_1=1$ and $\|\bm{d}_{-S}\|_1\le L$ we can get that it is bounded below by $\Lambda_{\min}(R) n^{-1}$. This gives us one possible value for the compatibility constant $\vt^2(L,S)$, notice that this includes the special case of traditional wavelet regression with $R = I$ and $\Lambda_{\min}(R) = 1$. 

Thus we have that 
\begin{equation}
	 n^{-1}\|\wh{\bm{f}} - \bm{f}^0\|_2^2 \le n^{-1}\|\bm{f}^0 - RW^{\top}\bm{d}^*\|_2^2 + \frac{9n|S|{\lambda^2_{\e}}}{\Lambda_{\min}(R)}.
\end{equation}

\textbf{Stochastic part}

We focus on obtaining a possible values for $\lambda_{\e}$. We start with the simple case where $R = I$ and $\bm{\e} \sim \mathcal{N}(0, \sigma^2I)$, i.e. the traditional wavelet approach with regularly spaced data. In this case we need to find a $\lambda_{\e}$ such that 
\begin{equation}
	\lambda_{\e} \ge \|W\bm{\e}\|_{\infty}/n.
\end{equation}
First note that $\bm{\e}^{\prime} = W\bm{\e}/\sigma \sim \mathcal{N}(0, I)$ by orthogonality of $W$. Hence we have  
\begin{equation}
	Pr\left( \|\bm{\e}^{\prime}\|_{\infty} > \sqrt{t^2+2\log n}\right) \le 2p\exp\left[ - \frac{t^2 + 2\log p}{2} \right] = 2\exp(-t^2/2).
\end{equation}
Thus with probability at-least $1-2\exp(-t^2/2)$ we have $\sigma\sqrt{t^2+2\log n}\ge \|W\bm{\e}\|_{\infty}$. Thus in this case we can take $\lambda_{\e} = n^{-1}\sigma\sqrt{t^2+2\log n}$. In the general case we would have the mean zero, sub-Gaussian $K$-vector $WR^{\top}\bm{\e}$. By a slightly more involved argument we can show that we can take $\lambda_{\e} = n^{-1}c_1\sqrt{t^2+2\log K}$ where $c_1$ depends on the distribution of $\e$ (i.e., the parameters of the sub-gaussian distribution) and matrix $R$. 

Thus we have shown so far that with probability at-least $1-2\exp(-t^2/2)$ we have
\begin{equation}
n^{-1}\|\wh{\bm{f}} - \bm{f}^0\|_2^2 \le n^{-1}\|\bm{f}^0 - RW^{\top}\bm{d}^*\|_2^2 + \frac{9c_1^2 }{\Lambda_{\min}(R)} \frac{|S|(t^2+2\log K)}{n},
\end{equation}
or without worrying about optimal constants we get the rate
\begin{equation}
n^{-1}\|\wh{\bm{f}} - \bm{f}^0\|_2^2 \le n^{-1}\|\bm{f}^0 - RW^{\top}\bm{d}^*\|_2^2 + C \frac{|S|\log K}{n}.
\end{equation}
To obtain our result we just need the final step: approximation error. 

\textbf{Approximation error part}

Now we will bound the term $n^{-1}\|\bm{f}^0 - RW^{\top}\bm{d}^*\|_2^2$. We will define specific types of vectors $\bm{d}^*$ which leads to specific sparse indes sets $S$. We begin with the decomposition:
\begin{equation}
	n^{-1}\|\bm{f}^0 - RW^{\top}\bm{d}^*\|_2^2 \le 2n^{-1}\|\bm{f}^0 - R\wt{\bm{f}}^0\|_2^2 + 2n^{-1}\| R\wt{\bm{f}}^0 - RW^{\top} \bm{d}^* \|_2^2,
\end{equation}
where $\wt{f}^0$ is the function obtained by interpolating $f^0$ from the data $(i/K, f^0(i/K))$ for $i=1,\ldots,K$ and $\wt{f}^0 = [\wt{f}^0(1/K), \ldots, \wt{f}^0(K/K)]^{\top}$.

% There are a number of results for interpolation error, e.g., a simple bound for piecewise linear interpolation over the grid $i/K$ ($i=1,\ldots,K$) with $|\frac{d^2}{dx^2} f^0(x)|\le C_1$  is
%\begin{align*}
%	\max_{x}|f^0(x) - \wt{f}^0(x)|\le \frac{C_1}{2K^2}.
%\end{align*}

For the second term, define $\Lambda_{\max}(R)$ as the maximum eigenvalue of $R^{\top}R$ then
\begin{align*}
	n^{-1}\| R\wt{\bm{f}}^0 - RW^{\top} \bm{d}^* \|_2^2 \le \Lambda_{\max}(R)n^{-1}\|\wt{\bm{f}}^0 - W^{\top}\bm{d}^*\|_2^2\le   \Lambda_{\max}(R)\|\wt{\bm{f}}^0 - W^{\top}\bm{d}^*\|_{\infty}^2. 
\end{align*}

For the last part we now define $\bm{d}^*$, the vector of wavelet coefficients such that it defines a function $f^*$ as a linear combination of wavelet basis functions. To be precise we have that
\begin{equation}
	f^*(x) =  \sum_{k=0}^{2^{j_0}-1}\phi_{j_0k}(x)\alpha^0_{j_0k} + \sum_{j=j_0}^{J^*-1}\sum_{k=0}^{2^j-1}\psi_{jk}(x)\beta^0_{jk},
\end{equation}
for some integer $J^*$, and where $\alpha_{j0k}^0$ and $\beta^0_{jk}$ are the wavelet coefficients of the true function $f^0$. 
Now we obtain:
\begin{align*}
	\max_{x} |f^*(x) - f^0(x)| &= \max_x \Big| \sum_{j=J^*}^{\infty}\sum_{k=0}^{2^j-1}\psi_{jk}(x)\beta^0_{jk} \Big|\\
	&\le \max_{x}\max_{j\ge J^*,k}|\psi_{jk}(x) |  \sum_{j=J^*}^{\infty}\sum_{k=0}^{2^j-1} |\beta^0_{jk}| \\
	&= \max_{x}\max_{j\ge J^*,k}|\psi_{jk}(x) |  \sum_{j=J^*}^{\infty}\|\bm{\beta}^0_j\|_1,
\end{align*}
where $\bm{\beta}_j\in \mathbb{R}^{2^j}$ is the mother wavelet coefficient vector at level $j$. Now assuming that $f^0\in B_{q_1,q_2}^s$
\begin{align*}
	\sum_{j=J^*}^{\infty}\|\bm{\beta}^0_j\|_1 &= 	\sum_{j=J^*}^{\infty}\frac{2^{js'}}{2^{js'}} \|\bm{\beta}^0_j\|_1, \qquad (s' = s-1/2)\\
	&\le  \left[ \sum_{j=J^*}^{\infty}\left( 2^{js'}\|\bm{\beta}^0_j\|_1 \right)^{q_2} \right]^{1/q_2}\left[ \sum_{j=J^*}^{\infty} 2^{-js'q_2'} \right]^{1/q'_2},
\end{align*}
where $q_2'$ is such that $1/q_2+1/q'_2=1$. Using the inequality $\|\bm{\beta}^0_j\|_1 \le 2^{j(1-1/q_1)} \|\bm{\beta}^0_j\|_{q_1}$ we get
\begin{align*}
\sum_{j=J^*}^{\infty}\|\bm{\beta}^0_j\|_1 &\le  \left[ \sum_{j=J^*}^{\infty}\left( 2^{j(s+1/2-1/q_1)}\|\bm{\beta}^0_j\|_{q_1} \right)^{q_2} \right]^{1/q_2}\left[ \sum_{j=J^*}^{\infty} 2^{-js'q_2'} \right]^{1/q'_2}\\
&= \left[ \sum_{j=J^*}^{\infty}\left( 2^{j(s+1/2-1/q_1)}\|\bm{\beta}^0_j\|_{q_1} \right)^{q_2} \right]^{1/q_2}\times C_22^{-J^*s},
\end{align*}
where the second term can be obtained by looking at $S_{\infty} - S_{J^*-1}$ where $S_n = \sum_{j=0}^{n}2^{-js'q_2'}$. The first term is bounded because $f^0\in B_{q_1,q_2}^s$.

\textbf{Putting the pieces together}

Thus we have shown so far, by taking $\bm{d}^*$ as defined above and $S$ being the active set of $\bm{d}^*$ (i.e. $|S| = 2^{J^*}$), that the rate is of the form (upto constants)
\begin{align*}
	n^{-1}\|\wh{\bm{f}} - \bm{f}^0\|_2^2 \le 2n^{-1}\|\bm{f}^0 - R\wt{\bm{f}}^0\|_2^2 + C_22^{-(2s)J^{*}} + C_3 2^{J^*}\frac{\log K}{n}.
\end{align*}
Treating the above as a function of $J^*$ and minimizing we obtain the approximate truncation order $|S| = \mathcal{O}(n^{1/(2s+1)})$ which minimizes the right hand side. Finally, putting all the different pieces together we obtain the bound: 
\begin{align*}
n^{-1}\|\wh{\bm{f}} - \bm{f}^0\|_2^2 \le C_4 \left( \frac{\log K}{n} \right)^{\frac{2s}{2s+1}} + 2n^{-1}\|\bm{f}^0 - R\wt{\bm{f}}^0\|_2^2.
\end{align*}

	% !TeX root = ../MANUSCRIPT.tex
\section{Proofs for additive \name} 

\subsection{Initial results}

We will present results in greater generality here. In the interest of brevity and clarity of exposition we avoided some technical details such as identifiablity and the intercept term in the model. We go into these details here. 

Let $f^*$ be a sparse additive approximation to $f^0$,  
\begin{equation*}
f^*(  {\bm{x}_i}) = c^0 + \sum_{j = 1}^p f^*_j(x_{ij}) = c^0 + \sum_{j \in S} f^*_j(x_{ij}),
\end{equation*}
where $S = \{j:f_j^*\not= 0\}$, which we call the active set, is a subset of $\{1,\ldots, p\}$ of size $|S|$ and, $c^0 = E(\bar{\bm{y}})$ where $\bar{\bm{y}}$ is the sample mean. To ensure identifiability, we assume $\sum_{i=1}^{n} f^*_j(x_{ij}) = 0$  $(j =1, \ldots, p).$

 We consider a large class of estimators of the type:
\begin{equation}
\wh{f}_1,\ldots, \wh{f}_p = \underset{(f_j)_{j=1}^p\in \mathcal{F}}{ \arg\min }\  \frac{1}{2n}\sum_{i=1}^{n} \Big\{ y_i - \bar{\bm{y}} - \sum_{j=1}^{p}f_{j}(x_{ij}) \Big\}^2 + \lambda_n\sum_{j=1}^{p}I(f_j)\ ,
\label{eqn:sparseAdditiveOptim}
\end{equation}
where $I(\cdot)$ is a penalty of the form 
%\begin{equation}
$I(f_j) = \|f_j\|_n + \lambda_n \up(f_j),$
%\label{eqn:penlatyForm}
%\end{equation}
for a semi-norm $\up(\cdot)$ and, empirical norm $\|\cdot\|_n$ defined for component $f_j$ as $\|f_j\|_n^2 = n^{-1}\sum_{i=1}^{n}[f_j(x_{ij})]^2$. In our case $\up(\cdot)$ is the Besov norm of the $B_{1,1}^s$ space. 

Throughout this proof, instead of the smoothness level $s$, we will use $\alpha = 1/s$. Before we begin the main proof, we define the notion of metric entropy which will be used throughout the proof.  For a set $\mathcal{F}$ equipped with some metric $d(\cdot,\, \cdot)$, the subset $\{f_1,\ldots,f_N\} \subset \mathcal{F}$ is a $\delta$-cover if for any $f\in \mathcal{F}$
%\begin{equation}
$
\min_{1\le i \le N} d(f,\, f_i) \le \de.
$
%\end{equation}
The log-cardinality of the smallest $\de$-cover is the $\de$-entropy of $\mathcal{F}$ with respect to metric $d(\cdot,\, \cdot)$. We denote by $H(\de,\, \mathcal{F},\, Q)$, the $\de$-entropy of a function class $\mathcal{F}$ with respect to the $\|\cdot\|_Q$ metric for a measure $Q$, where $\|f\|_Q^2 =\int \{f(x)\}^2\,dQ(x)$.  For a fixed sample of covariate $j$, $x_{1j},\ldots,x_{nj}$, we denote by $Q_{nj}$ the empirical measure $Q_{nj} = n^{-1}\sum_{i=1}^{n}\delta_{x_{ij}}$ and use the short-hand notation $\|\cdot\|_n = \|\cdot\|_{Q_{nj}}$.

The main ingridient we require for proving results for sparse additive models is the entropy condition, specifically we require 
\[H(\delta, \{f_j\in \mathcal{F} : \up(f_j)\le 1\}, Q_{nj}) \le A_0\delta^{-\alpha},\] 
for $\alpha\in (0,2)$, and so forth. 

\textbf{Note:} In the case of our Besov norm, the above entropy condition holds for $\alpha = 1/s$, i.e., 
\begin{align*}
	H(\delta, \{f_j\in \mathcal{F} : \up(f_j)\le 1\}, Q_{nj}) \le A_0\delta^{-1/s}.
\end{align*}

\begin{lemma}[Basic inequality]
	For any function $f^* = \sum_{j=1}^{p}f_j^*$, where $f_j^*\in \mathcal{F}$ and, the solution $\wh{f}$ of (\ref{eqn:sparseAdditiveOptim}), we have the following basic inequality
	\begin{equation*}
		\frac{1}{2}\|\wh{f} - f^0\|_n^2 + \lambda I_p(\wh{f}) \le  |\langle \e,\, \wh{f} - f^* \rangle_n| + \lambda I_p(f^*) + |\bar{\e}|\sum_{j=1}^{p}\|\wh{f}_j - f^*_j\|_n  +  \frac{1}{2}\|f^* - f^0\|_n^2,
	\end{equation*}
	where $\langle \e, f\rangle_n = \frac{1}{n}\sum_{i=1}^{n}\e_if({\bm{x}_i})$, $\bar{\e} = \frac{1}{n}\sum_{i=1}^{n}\e_i$ and $I_p(f) = \sum_{j=1}^{p}I(f_j) = \sum_{j=1}^{p}\|f_j\|_n + \lambda\up(f_j)$ for an additive function $f$.
\end{lemma}

\begin{proof}
	We have 
	\begin{align*}
		\frac{1}{2n}\sum_{i=1}^{n} \left\{ y_i - \bar{\bm{y}} - \wh{f}({\bm{x}_i}) \right\}^2 + \lambda I_p(\wh{f}) \le \frac{1}{2n}\sum_{i=1}^{n} \left\{ y_i - \bar{\bm{y}} - f^*({\bm{x}_i}) \right\}^2 + \lambda I_p(f^*_j),
	\end{align*}
	\begin{align*}
		&\Leftrightarrow \frac{1}{2n}\sum_{i=1}^{n} \left\{ \e_i + c^0 - \bar{\bm{y}} -(\wh{f} - f^0)({\bm{x}_i})  \right\}^2 + \lambda I_p(\wh{f}) \le \frac{1}{2n}\sum_{i=1}^{n} \left\{ \e_i + c^0 - \bar{\bm{y}}-(f^*-f^0)({\bm{x}_i})  \right\}^2 + \lambda I_p(f^*_j)\\
	\end{align*}
	\begin{align*}
		&\Rightarrow \frac{1}{2n}\sum_{i=1}^{n} \left( \e_i + c^0 - \bar{\bm{y}}\right)^2 +(\wh{f} - f^0)^2({\bm{x}_i}) -2(\e_i+c^0-\bar{\bm{y}})(\wh{f} - f^0)({\bm{x}_i}) + \lambda I_p(\wh{f}) \\
		&\le \frac{1}{2n}\sum_{i=1}^{n} \left( \e_i+c^0 - \bar{\bm{y}}  \right)^2 + (f^*-f^0)^2({\bm{x}_i})-2(\e_i+c^0-\bar{\bm{y}})(f^*-f^0)({\bm{x}_i}) + \lambda I_p(f^*)\\
		&\Rightarrow \frac{1}{2}\|\wh{f} - f^0\|_n^2 -\langle \e + c^0 -\bar{\bm{y}},\, \wh{f}-f^0\rangle_n + \lambda I_p(\wh{f}) \\
		&\le   \frac{1}{2}\|f^*-f^0\|_n^2- \langle \e+c^0-\bar{\bm{y}},\, f^*-\wh{f}+\wh{f}-f^0 \rangle_n +\lambda I_p(f^*)\\
		&\Rightarrow \frac{1}{2}\|\wh{f} - f^0\|_n^2 - \langle \e + c^0 - \bar{\bm{y}},\, \wh{f}-f^0\rangle_n + \lambda I_p(\wh{f}) \\
		&\le   \frac{1}{2}\|f^*-f^0\|_n^2 - \langle \e + c^0 - \bar{\bm{y}},\, f^*-\wh{f}\rangle_n - \langle \e+c^0-\bar{\bm{y}},\,\wh{f}-f^0 \rangle_n +\lambda I_p(f^*),
	\end{align*}
	which implies
	\begin{align*}
		&\frac{1}{2}\|\wh{f} - f^0\|_n^2  + \lambda I_p(\wh{f}) \le   \frac{1}{2}\|f^*-f^0\|_n^2 - \langle \e + c^0 - \bar{\bm{y}},\, f^*-\wh{f}\rangle_n  +\lambda I_p(f^*) \\
		\Rightarrow&\frac{1}{2}\|\wh{f} - f^0\|_n^2  + \lambda I_p(\wh{f}) \le  |\langle \e,\, \wh{f} - f^* \rangle_n|+ \sum_{j=1}^{p}\langle c^0-\bar{\bm{y}},\wh{f}_j - f^*_j \rangle_n +\lambda I_p(f^*) +  \frac{1}{2}\|f^* - f^0\|_n^2\\
		\Rightarrow&\frac{1}{2}\|\wh{f} - f^0\|_n^2  + \lambda I_p(\wh{f}) \le  |\langle \e,\, \wh{f} - f^* \rangle_n|+ | c^0-\bar{\bm{y}}| \sum_{j=1}^{p}\|\wh{f}_j - f^*_j\|_n +\lambda I_p(f^*) +  \frac{1}{2}\|f^* - f^0\|_n^2.
	\end{align*}
	Now for the second term note that:
	\begin{align*}
		|c^0 - \bar{\bm{y}}| &= \left|\frac{1}{n}\sum_{i=1}^{n}(c^0 - y_i)\right| = \left| \frac{1}{n}\sum_{i=1}^{n}\left\{ c^0 - c^0 -\sum_{j=1}^{p}f_j^0(x_{i,j}) -\e_i\right\}\right| = |\bar{\e}|.
	\end{align*}
	Which leads us to 
	\begin{equation*}
		\frac{1}{2}\|\wh{f} - f^0\|_n^2 + \lambda I_p(\wh{f}) \le  |\langle \e,\, \wh{f} - f^* \rangle_n| + \lambda I_p(f^*) + |\bar{\e}|\sum_{j=1}^{p}\|\wh{f}_j - f^*_j\|_n  +  \frac{1}{2}\|f^* - f^0\|_n^2.
	\end{equation*}
\end{proof}

\begin{lemma}[Bounding the term $|\bar{\e}|$]
	For $\e = (\e_1,\ldots\e_n)^T$ such that $E(\e_i) = 0$ and
	\begin{equation*}
		L^2\left\{ {E}\Big(e^{\e_i^2/L^2}\Big) -1 \right\}\le \sigma_0^2\ ,
	\end{equation*}
	for all $\kappa>0$ and
	\begin{equation*}
		\rho = \kappa\max\left\{ n^{-\frac{1}{2+\alpha}} , \left( \frac{\log p}{n} \right)^{1/2} \right\},
	\end{equation*}
	we have that with probability at least $1- 2\exp\left( -{n\rho^2}/{c_1} \right)$,
	\begin{equation*}
		|\bar{\e}|\le \rho ,
	\end{equation*}
	for a constant $c_1$ that depends on $L$ and $\sigma_0$.
\end{lemma}
\begin{proof}
	By Lemma 8$\cdot$2 of \cite{vandegeer2000empirical}~(with $\gamma_n = {1}_n/n$) we have for all $t> 0$
	\begin{align*}
		\text{pr}\left( \left| \frac{1}{n}\sum_{i=1}^{n}\e_i \right| \ge t \right) \le 2\exp\left\{ -\frac{nt^2}{8(L^2+\sigma_0^2)} \right\}.
	\end{align*}
	The result follows by setting $t = \rho$.
\end{proof}

%\textbf{Recall Notation:} $\hd_j = \wh{f}_j-f^0_j$ and $H(\delta, \mathcal{F}, Q)$ is the $\delta$ entropy of the function class $\mathcal{F}$ with respect to the $L^2(Q)$ norm for some measure $Q$ We also define the empirical measure $Q_n$ such that $\|f\|^2_{Q_n} = \|f\|^2_n = \frac{1}{n}\sum_{i=1}^{n}[f(\bm{x}_i)]^2$.

\begin{lemma}[Bounding the term $|\langle{\e}, \wh{f} - f^*\rangle_n| $]
	For $\lambda\ge 4\rho$ where $$\rho = \kappa\max\left\{ n^{-\frac{1}{2+\alpha}} , \left( \frac{\log p}{n} \right)^{1/2} \right\},$$ for some constant $\kappa$, if
	\begin{equation*}
		H(\delta, \{f\in \mathcal{F}:\up(f)\le 1 \}, Q_n) \le A_0\delta^{-\alpha},
	\end{equation*} 
	we then have with probability at least $1-c_2\exp\left(-c_3n\rho^2\right)$
	\begin{equation*}
		|\langle \e,\wh{f}_j - f^*_j\rangle_n|\le \rho \|\wh{f}_j - f^*_j\|_n + \rho\lambda \up(\wh{f}_j - f^*_j),
	\end{equation*}
	for all $j=1,\ldots,p$ and positive constants $c_2$ and $c_3$.
\end{lemma}

\begin{proof}
	Firstly, for $\mathcal{F}_0 = \{f\in \mathcal{F}: \up(f)\le 1\}$ we have by assumption a $\delta$ cover $f_1,\ldots,f_N$ such that for all $f\in \mathcal{F}_0$ we have $
	\min_{j\in \{1,\ldots,N\}} \| f_j -  f\|_n\le \delta$. Now we are interested in the set $\mathcal{F}_{0,\lambda} = \{f\in \mathcal{F}: \lambda\up(f)\le 1\}$.
	%Firstly, we note that $\{\lambda f : f\in \mathcal{F}\text{ and } \up(f) \le 1\} \subset \mathcal{F}_{0,\lambda}$. 
	Firstly, for a function $f\in \mathcal{F}_{0,\lambda}$,
	\begin{align*}
		\min_{j\in \{1,\ldots,N\}} \| f - f_j/\lambda\|_n = \min_{j\in \{1,\ldots,N\}} \frac{1}{\lambda}\| \lambda f - f_j\|_n\le  \frac{\delta}{\lambda},
	\end{align*}
	because $\up(\lambda f) = \lambda \up(f) \le 1\Rightarrow  \lambda f \in \mathcal{F}_0$. This means that the set $\{f_1/\lambda,\ldots, f_N/\lambda\}$ is a $\de/\lambda$ cover of the set $\mathcal{F}_{0,\lambda}$.
	
	This implies that $H(\delta, \mathcal{F}_0,Q_n) \le A_0\delta^{-\alpha}\Rightarrow H(\delta/\lambda, \mathcal{F}_{0,\lambda},Q_n) \le A_0\delta^{-\alpha}$ or equivalently $H(\delta, \mathcal{F}_{0,\lambda}, Q_n) \le A_0(\delta\lambda)^{-\alpha}$. Finally, since $\{f\in\mathcal{F}: I(f)\le 1\} \subset \{f\in\mathcal{F}: \up(f)\le \lambda^{-1}\}$ we have 
	\begin{align*}
		H(\delta, \{f\in \mathcal{F}: I(f)\le 1 \}, Q_n) \le A_0(\delta\lambda)^{-\alpha} .
		%\le A_1(\delta\rho)^{-\alpha},
	\end{align*}
	%since $\lambda^{-1}\le \rho^{-1}/4$.

	The same entropy bound holds for the class 
	\begin{equation}
		\label{eqn:class-scaled2}
		\wt{\mathcal{F}} = \left\{\frac{f_j - f_j^*}{\|f_j - f_j^*\|_n + \lambda \Om(f_j - f_j^*)}: f_j\in \mathcal{F}\right\},
	\end{equation}
	and we can now apply Corollary 8.3 of \cite{vandegeer2000empirical} by noting that 
	\begin{align*}
		\int_{0}^{1} H^{1/2}(u, \wt{\mathcal{F}}, Q_n)\, du \le \wt{A}_0\lambda^{-\alpha/2},
	\end{align*}
	for some constant $\wt{A}_0 = \wt{A}_0(A_0)$. For some $c_2 = c_2(L,\sigma_0)$ and all
	$\delta\ge 2c_2\wt{A}_0\lambda^{-\alpha/2}n^{-1/2}$ 
	we have 
	\begin{align}
		\label{eqn:lemmaUse}
		\text{pr}\left( \sup_{f_j\in \mathcal{F}}  \frac{\left| \langle \e, {f}_j - f^*_j \rangle_n\right|}{ \|f_j-f_j^*\|_n + \lambda \Om(f_j-f_j^*) } \ge \delta \right) \le c_2\exp\left( -\frac{n\delta^2}{4c_2^2} \right).
	\end{align}
	Since $\lambda \ge \rho$ we note that $2c_2\wt{A}_0\lambda^{-\alpha/2}n^{-1/2} \le 2c_2\wt{A}_0\rho^{-\alpha/2}n^{-1/2}$ and that 
	\begin{align*}
		2c_2\wt{A}_0\rho^{-\alpha/2}n^{-1/2} \le \rho \Leftrightarrow \rho \ge \left( 2c_2\wt{A}_0 \right)^{\frac{2}{2+\alpha}} n^{-\frac{1}{2+\alpha}}.
	\end{align*}
	Which holds by definition since $\rho = \kappa\max\left\{ \left({\log p}/{n}\right)^{1/2}, n^{-{1}/{(2+ \alpha)}} \right\} \ge \kappa n^{-{1}/{(2+\alpha)}}$ and $\kappa$ is sufficiently large
	(any $ \kappa \ge \Big( 2c_2\wt{A}_0 \Big)^{{2}/{(2+\alpha)}}$ would suffice). Therefore, we can take $\delta = \rho$ in \eqref{eqn:lemmaUse} along with a union bound to obtain 
	\begin{align*}
		\text{pr}\left( \max_{j=1,\ldots,p}\sup_{f_j\in \mathcal{F}}  \frac{\left| \langle \e, {f}_j - f^*_j \rangle_n\right|}{ \|f_j-f_j^*\|_n + \lambda \Om(f_j-f_j^*) } \ge {\rho} \right) &\le pc_2\exp\left( - \frac{n\rho^2}{4c_2^2} \right)\\
		&= c_2\exp\left\{ - n\rho^2\left( \frac{1}{4c_2^2} - \frac{\log p}{n\rho^2} \right)\right\}\\
		&\le c_2\exp\left( - n\rho^2c_3 \right),
	\end{align*}
	for some positive constant $c_3 = c_3(c_2,\wt{A}_0)$.
	
	% % % % % % % % % % % % % % % % % % % % % % % % %
	% % % % % % % % % % % % % % % % % % % % % % % % %
	% % % % % % % % % % % % % % % % % % % % % % % % %
	% % % % % % % % % % % % % % % % % % % % % % % % %

	Finally, we show that $c_3> 0$. This follows from the fact that $ {1}/{(4c_2^2)} - {\log p }/{(n\rho^2)}> 0\Leftrightarrow n\rho^2 > 4c_2^2\log p$. This holds since $n\rho^2\ge \kappa^2\log p$ for $\kappa$ sufficiently large. Thus, we have with probability at least $1-c_2\exp\left(c_3n\rho^2\right)$ for all $j= 1,\ldots,p$
	\begin{align*}
		|\langle\e,\wh{f}_j - f^*_j\rangle_n| \equiv |\langle\e,\hd_j\rangle_n| \le \rho \|\hd_j\|_n + \rho\lambda\up(\hd_j)\ .
	\end{align*}
\end{proof}
\subsection{Using the active set}
So far we have shown that, for $\lambda\ge 4\rho$, with probability at least $1-2\exp\left(-{n\rho}/{c_1}\right) -c_2\exp\left(-c_3n\rho^2\right)$, the following inequality holds
\begin{align*}
	\|\wh{f} - f^0\|_n^2 + 2\lambda\sum_{j=1}^{p}I(\wh{f}_j) &\le 2|\langle \e, \wh{f}-f^* \rangle_n| + 2|\bar{\e}|\sum_{j=1}^p \|\hd_j\|_n + 2\lambda \sum_{j=1}^{p}I(f^*_j)+\|f^*-f^0\|_n^2 \\
	&\le \left\{ \sum_{j=1}^{p} 2\rho\|\hd_j\|_n + 2\rho\lambda\up(\hd_j)\right\}  + \left( 2\rho \sum_{j=1}^{p}\|\hd_j\|_n \right) \\
	&+ \left\{ 2\lambda \sum_{j=1}^{p}I(f^*_j)\right\}+\|f^*-f^0\|_n^2\\
	\Rightarrow \|\wh{f}-f^0\|_n^2 + 2\lambda\sum_{j=1}^{p}I(\wh{f}_j)  &\le \sum_{j=1}^{p} \left\{ \lambda \|\hd_j\|_n + \frac{\lambda^2}{2} \up(\hd_j) + 2\lambda\|f^*_j\|_n+2\lambda^2\up(f^*_j)\right\} + \|f^* - f^0\|_n^2.
\end{align*}
For notational convenience we will exclude the $\|f^*-f^0\|_n^2$ term in the following manipulations. If $S$ is the active set then we have on the right hand side,
\begin{align*}
	\mathrm{RHS} &= \lambda \sum_{j\in S} \left\{\|\hd_j\|_n +  \frac{\lambda}{2}\up(\hd_j) +2\|f^*_j\|_n + 2\lambda\up(f^*_j) \right\}+\lambda\sum_{j\in S^c} \left\{\|\wh{f}_j\|_n+\frac{\lambda}{2}\up(\wh{f}_j)\right\}\\
	&\le \lambda \sum_{j\in S} \left\{\|\hd_j\|_n +  \frac{\lambda}{2}\up(\hd_j) +2\|\hd_j\|_n+2\|\wh{f}_j\|_n + 2\lambda\up(f^*_j) \right\}+\lambda\sum_{j\in S^c} \left\{\|\wh{f}_j\|_n+\frac{\lambda}{2}\up(\wh{f}_j)\right\}\\
	&= 3\sum_{j\in S}\lambda\|\hd_j\|_n+2\sum_{j\in S}\lambda^2\up(f^*_j)+ \sum_{j\in S^c}\lambda\|\wh{f}_j\|+\frac{1}{2}\sum_{j\in S^c}\lambda^2\up(\wh{f}_j) + 2\sum_{j\in S}\lambda\|\wh{f}_j\|_n + \frac{1}{2}\sum_{j\in S}\lambda^2\up(\hd_j),
\end{align*}
where the inequality holds by the decomposition $\|f^*_j\|_n = \|f^*_j-\wh{f}_j+\wh{f}_j\|_n\le \|\hd_j\|_n + \|\wh{f}_j\|_n$.

\noindent On the left hand side we have
\begin{align*}
	\mathrm{LHS} &= \|\wh{f} - f^0\|_n^2+2\lambda\sum_{j\in S} \left\{\|\wh{f}_j\|_n+\lambda\up(\wh{f}_j)\right\} + 2\lambda\sum_{j\in S^c} \left\{\|\wh{f}_j\|_n+\lambda\up(\wh{f}_j)\right\}\\
	&\ge \|\wh{f} - f^0\|_n^2+2\lambda\sum_{j\in S} \left\{\|\wh{f}_j\|_n +\lambda\up(\hd_j)-\lambda\up(f^*_j)\right\} + 2\lambda\sum_{j\in S^c} \left\{\|\wh{f}_j\|_n+\lambda\up(\wh{f}_j)\right\},
\end{align*}
where the inequality follows from the triangle inequality $  \up(\wh{f}_j) + \up(f^*_j)\ge \up(\hd_j)$ since $\up(\cdot)$ is a semi-norm. By re-arranging the terms we obtain the inequality
\begin{equation*}
	%\small
	\|\wh{f} - f^0\|_n^2 + \lambda\sum_{j\in S^c}\left\{ \|\wh{f}_j\|_n + \frac{3\lambda}{2}\up(\wh{f}_j) \right\} + \frac{3\lambda^2}{2}\sum_{j\in S}\up(\hd_j) \le 3 \lambda\sum_{j\in S} \|\hd_j\|_n + {4\lambda^2} \sum_{j\in S}\up(f^*_j) + \|f^*-f^0\|_n^2
\end{equation*}
which implies that 
\begin{equation*}
	\label{eqn:inequalitySlowRates2}
	\|\wh{f} - f^0\|_n^2 + \lambda\sum_{j\in S^c} \|\hd_j\|_n + \frac{3\lambda^2}{2}\sum_{j=1}^p\up(\hd_j) \le 3 \lambda\sum_{j\in S} \|\hd_j\|_n + {4\lambda^2} \sum_{j\in S}\up(f^*_j) + \|f^*-f^0\|_n^2.
\end{equation*}
This implies the slow rates for convergence for $\lambda \ge 4\rho$ and $|S|$
\begin{equation*}
	\label{eqn:inequalitySlowRates3}
	\frac{1}{2}\|\wh{f} - f^0\|_n^2 + \le |S|\lambda\left\{ 3\sum_{j\in S} \|\hd_j\|_n/|S| + 2\lambda\sum_{j\in S}\up(f^*_j)/|S|\right\}  + \frac{1}{2}\|f^*-f^0\|_n^2.
\end{equation*}

This completes the proof of the first part of the theorem. Recall that $\lambda$ is of the order:
\[\kappa\max\left\{ n^{-\frac{1}{2+\alpha}} , \left( \frac{\log p}{n} \right)^{1/2} \right\},\]
and for the Besov space $B_{1,1}^s$ we have $\alpha = 1/s$.

\subsection{Using the compatibility condition}
\label{sec:compatibility}

Recall the compatibility condition for $f = \sum_{j=1}^{p}f_j$, whenever
\begin{equation}
	\label{eqn:compatibility1}
	\sum_{j\in S^c}\|f_j\|_n + \frac{3\lambda}{2}\sum_{j=1}^p\up(f_j) 
	\le 3\sum_{j\in S} \|f_j\|_n,
\end{equation}
then we have 
\begin{equation*}
	\sum_{j\in S}\|f_j\|_n \le |S|^{1/2}\|f\|_n/\vartheta(S).
\end{equation*}
Once we assume the compatibility condition we can prove the rest of the theorem by considering the following two cases.

\noindent \textbf{Case 1:} $\lambda\sum_{j\in S}\|\hd_j\|_n\ge 4\lambda^2\sum_{j\in S}\up(f^*_j)$ in which case we have
\begin{align*}
	\|\wh{f} - f^0\|_n^2 + \lambda\sum_{j\in S^c} \|\hd_j\|_n + \frac{3\lambda^2}{2}\sum_{j=1}^p\up(\hd_j) &\le 4 \lambda \sum_{j\in S} \|\hd_j\|_n + \|f^* - f^0\|_n^2 \ ,
\end{align*}
hence for the function $\wh{f} - f^* = \sum_{j=1}^{p}{\hd}_j$ (\ref{eqn:compatibility1}) holds and hence by the compatibility condition we have
\begin{align*}
	\|\wh{f} - f^0\|_n^2 &+ \lambda\sum_{j\in S^c} \|\hd_j\|_n + \frac{3\lambda^2}{2}\sum_{j=1}^p\up(\hd_j) \le \frac{4\lambda |S|^{1/2}}{\vartheta(S)}\|\wh{f} - f^*\|_n + \|f^* - f^0\|_n^2 \\
	&\le \frac{4\lambda |S|^{1/2}}{\vartheta(S)}\|\wh{f} - f^0\|_n + \frac{4\lambda |S|^{1/2}}{\vartheta(S)}\|{f}^* - f^0\|_n + \|f^* - f^0\|_n^2\\
	&\le 2\left\{ \frac{2\lambda (2s)^{1/2}}{\vartheta(S)} \right\}\left( \frac{\|\wh{f} - f^0\|_n}{2^{1/2}} \right) + 2\left\{\frac{2\lambda |S|^{1/2}}{\vartheta(S)} \right\} \left(\|{f}^* - f^0\|_n\right) + \|f^* - f^0\|_n^2\\
	&\le  \frac{4\lambda^2 (2|S|) }{\vartheta^2(S)} + \frac{\|\wh{f} - f^0\|^2_n}{2}  + \frac{4\lambda^2 {|S|}}{\vartheta^2(S)} + \|{f}^* - f^0\|^2_n + \|f^* - f^0\|_n^2\\
	&\le \frac{12\lambda^2 |S| }{\vartheta^2(S)}+\frac{\|\wh{f} - f^0\|^2_n}{2}  +2\|{f}^* - f^0\|^2_n,
\end{align*}
where we use the inequality $2ab\le a^2+b^2$ and this implies that 
\begin{equation*}
	\frac{1}{2} \|\wh{f} - f^0\|_n^2 + \lambda\sum_{j\in S^c} \|\hd_j\|_n + \frac{3\lambda^2}{2}\sum_{j=1}^p\up(\hd_j) \le \frac{12s\lambda^2}{\vartheta^2(S)} + 2\|f^* - f^0\|_n^2.
\end{equation*}

\noindent \textbf{Case 2:} $\lambda\sum_{j\in S}\|\hd_j\|_n\le 4\lambda^2\sum_{j\in S}\up(f^*_j)$ in which case we have 
\begin{align*}
	\|\wh{f} - f^0\|_n^2 + \lambda\sum_{j\in S^c} \|\hd_j\|_n + \frac{3\lambda^2}{2}\sum_{j=1}^p\up(\hd_j) &\le 16 \lambda^2\sum_{j\in S} \up(f^*_j) + \|f^*-f^0\|_n^2 \\
	&\le 16|S|\lambda^2\sum_{j\in S}\up(f^*_j)/|S| + \|f^*-f^0\|_n^2,
\end{align*}
which implies
\begin{align*}
	\frac{1}{2}\|\wh{f} - f^0\|_n^2 + \lambda\sum_{j\in S^c} \|\hd_j\|_n + \frac{3\lambda^2}{2}\sum_{j\in S}\up(\hd_j) 
	&\le 16|S|\lambda^2\sum_{j\in S} \up(f^*_j)/|S| + 2\|f^*-f^0\|_n^2.
\end{align*}

	\bibliographystyle{plainnat}
	\bibliography{bibfileah/refnew}